%% file: main.tex
\documentclass[conference]{IEEEtran}
\IEEEoverridecommandlockouts
\usepackage{cite}
\usepackage{amsmath,amssymb,amsfonts}
\usepackage{algorithmic}
\usepackage{graphicx}
\usepackage{textcomp}
\usepackage{xcolor}
\usepackage{amsmath}
\usepackage{subcaption}
\usepackage{hyperref}
\usepackage{stackengine}
\usepackage{float}
\renewcommand\sin[1]{s{#1}}
\renewcommand\cos[1]{c{#1}}
\renewcommand\tan[1]{t{#1}}

\newcommand{\linebreakand}{%
  \end{@IEEEauthorhalign}
  \hfill\mbox{}\par
  \mbox{}\hfill\begin{@IEEEauthorhalign}
}
 
\allowdisplaybreaks
\def\BibTeX{{\rm B\kern-.05em{\sc i\kern-.025em b}\kern-.08em
    T\kern-.1667em\lower.7ex\hbox{E}\kern-.125emX}}
\begin{document}

\title{
An Efficient Detection and Control System for Underwater Docking using Machine Learning and Realistic Simulation: A Comprehensive Approach
\thanks{*This work is partially supported by ONR N00014-20-1-2085.}
\thanks{$^{1}$ J. Chavez-Galaviz, J. Li, M. Bergman, M. Mengdibayev are Graduate Research Assistants with the School of Mechanical Engineering, Purdue University, West Lafayette, IN, USA. {\tt\small}and
$^{2}$ N. Mahmoudian is an associate professor
        {\tt\small ninam@purdue.edu} with the school of Mechanical Engineering, Purdue University,
        West Lafayette, IN 47907, USA}
}

\author{
\IEEEauthorblockN{Jalil Chavez-Galaviz $^{1}$}
\IEEEauthorblockA{\textit{Mechanical Engineering} \\
\textit{Purdue University}\\
West Lafayette, Indiana \\
jchavezg@purdue.edu}
\and
\IEEEauthorblockN{Jianwen Li $^{1}$}
\IEEEauthorblockA{\textit{Mechanical Engineering} \\
\textit{Purdue University}\\
West Lafayette, Indiana \\
li3602@purdue.edu}
\and
\IEEEauthorblockN{Matthew Bergman $^{1}$}
\IEEEauthorblockA{\textit{Mechanical Engineering} \\
\textit{Purdue University}\\
West Lafayette, Indiana \\
bergman9@purdue.edu}
\and
\IEEEauthorblockN{Miras Mengdibayev $^{1}$}
\IEEEauthorblockA{\textit{Mechanical Engineering} \\
\textit{Purdue University}\\
West Lafayette, Indiana \\
mmengdib@purdue.edu}
\and
\IEEEauthorblockN{\hspace{20em}Nina Mahmoudian $^{2}$}
\IEEEauthorblockA{\hspace{22em}\textit{Mechanical Engineering} \\
\textit{\hspace{22em}Purdue University}\\
\hspace{22em}West Lafayette, Indiana \\
\hspace{22em}ninam@purdue.edu}

}

\maketitle

\begin{abstract}
Underwater docking is critical to enable the persistent operation of Autonomous Underwater Vehicles (AUVs). For this, the AUV must be capable of detecting and localizing the docking station, which is complex due to the highly dynamic undersea environment. Image-based solutions offer a high acquisition rate and versatile alternative to adapt to this environment; however, the underwater environment presents challenges such as low visibility, high turbidity, and distortion. In addition to this, field experiments to validate underwater docking capabilities can be costly and dangerous due to the specialized equipment and safety considerations required to conduct the experiments. This work compares different deep-learning architectures to perform underwater docking detection and classification. The architecture with the best performance is then compressed using knowledge distillation under the teacher-student paradigm to reduce the network's memory footprint, allowing real-time implementation. To reduce the simulation-to-reality gap, a Generative Adversarial Network (GAN) is used to do image-to-image translation, converting the Gazebo simulation image into a realistic underwater-looking image. The obtained image is then processed using an underwater image formation model to simulate image attenuation over distance under different water types. The proposed method is finally evaluated according to the AUV docking success rate and compared with classical vision methods. The simulation results show an improvement of 20\% in the high turbidity scenarios regardless of the underwater currents. Furthermore, we show the performance of the proposed approach by showing experimental results on the off-the-shelf AUV Iver3.

\end{abstract}


\section{Introduction}\label{sec:intro}

The movement toward ensuring more sustainable use of ocean resources has been growing along with the need for reducing environmental and human stressors on the planet to support a Networked Blue Economy.  The Blue Economy requires innovation toward more sustainable use of marine and coastal resources.  There is a unique opportunity to enhance existing ocean observation capabilities that quantify the responses of marine and coastal resources to climate change and other stressors, such as nutrient pollution and hypoxia, so that we can make informed and coordinated decisions about how to use marine resources sustainably. Enabling the persistent operation of maritime systems for a variety of applications in environmental studies will contribute to the implementation of Sustainable Development Goal 14 (SDG 14)\cite{lee2020blue,virto2018preliminary} to “Conserve and sustainably use the oceans, seas and marine resources for sustainable development.”

\begin{figure}[t]
\footnotesize
\centering
\stackunder[5pt]{\includegraphics[width=0.23\textwidth]{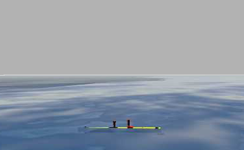}}{(a)}%
\hspace{0.005\textwidth}
\stackunder[5pt]{\includegraphics[width=0.23\textwidth]{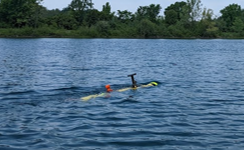}}{(b)}%
\hspace{0.005\textwidth}
\stackunder[5pt]{\includegraphics[width=0.49\textwidth]{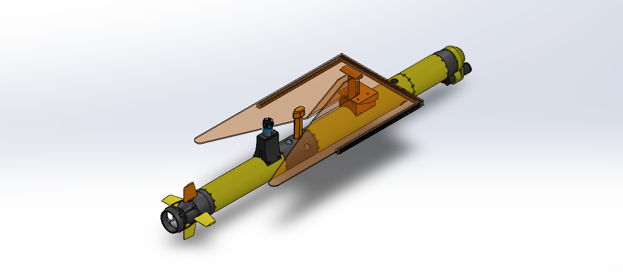}}{(c)}%

\caption{Underwater docking setup used in this work consisting of the AUV platform (Iver3), and the dock (flat funnel-shaped docking station) with a light attached to aid with the optical based underwater. (a) shows an image from the AUV going towards the docking station in simulation. (b) depicts the AUV approaching the docking station during field experiments. (c) illustrates a CAD model of the AUV docked in the docking station.}
\label{fig:intro}
\end{figure}

Persistent autonomy of Autonomous Underwater Vehicles (AUVs) is dependent on how to effectively respond to energy needs in the presence of dynamic conditions and how to account for substantial environmental uncertainty.  
Currently, most AUVs rely on batteries for operation \cite{petillot2019underwater}, and thus, they need to be recovered for recharge. Underwater docking, which is the process of connecting an AUV to a platform called a dock \cite{dhanak2016springer} can overcome AUVs' limitations since it allows a mechanism to recharge, recover, and transfer data during a mission, increasing the efficiency of underwater missions. As a result, underwater docking enables long-term long-distance sampling and data gathering operations with fewer resources. Fig.~\ref{fig:intro} shows the underwater docking setup in field experiments and a simulation environment. 

\begin{figure}[t]
    \centering
    \includegraphics[width=\columnwidth]{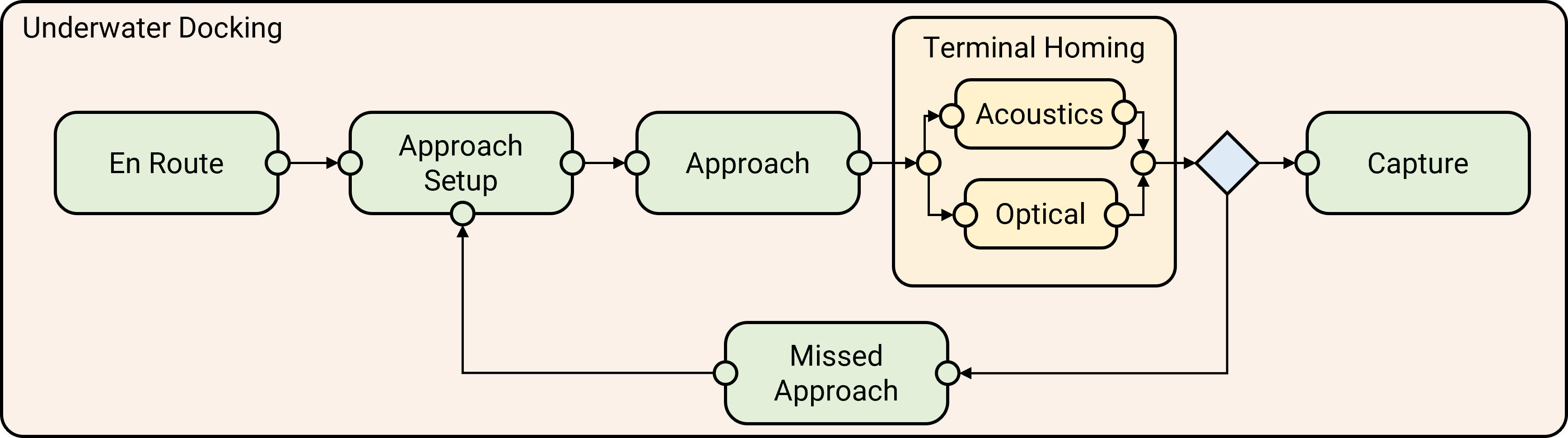}
    \caption{Multiple stages of the docking system, during the approach stages, the AUV gets close to the docking stations. During the terminal homing it uses a combination of optical and acoustic guidance to quickly adapt to the docking station position.}
    \label{fig:uwd_states}
\end{figure}

Achieving underwater docking requires a complex set of maneuvers or steps, which can vary depending on the specific design of the AUV and the docking station \cite{yazdani2020survey}; however, some general steps can be identified during the docking procedure, such as the approach and terminal homing  \cite{bellingham2016autonomous}. This work uses acoustic and vision sensing to bring AUV from the far-field to complete docking for data/energy transfer~\cite{page2021underwater}. First, the AUV approaches the dock using long-range navigation, such as dead reckoning. Once the AUV is within acoustic range using Ultra-Short Baseline (USBL), the AUV follows the approach trajectory to the docking station. For this stage, acoustics communication is a widely used method \cite{keane2022expediting,lin2022docking}, since it can work in a long range. When near the docking station, the AUV enters the terminal homing stage, in which it attempts to further improve its docking accuracy by using optical methods. This multi-stage process is illustrated in Fig.~\ref{fig:uwd_states}.

Optical guidance can be used during the terminal homing stage of the docking procedure since it offers an accurate and high update rate alternative \cite{vvekanandan2023autonomous,lin2022docking,page2021underwater,liu2018learning,li2018underwater,zhang2016auv}. Additionally, visual guidance, specifically cameras, can serve multiple purposes other than underwater docking, such as wildlife classification \cite{katija2022fathomnet}, which makes them a more cost-effective sensor to carry onboard. However, the undersea domain poses challenges such as a limited visibility range mainly due to the absorption and scattering effects \cite{liu2018learning}, especially in coastal areas where the water presents higher turbidity \cite{akkaynak2018revised}, and reflections or other light sources that can result in false detections of the docking station and potentially in unsuccessful docking attempts.

In recent years object detection and localization have benefited from the growth of deep learning. In this field, Convolutional Neural Networks (CNN) have revolutionized computer vision due to their ability to learn meaningful representations of images through feature extraction \cite{goodfellow2016deep,arroyo2016fusion}. While CNNs have been widely used for object localization and classification in aerial or terrestrial applications \cite{liu2023review,hsu2018vehicle}, they have not had the same impact on the underwater domain. Some of the existing underwater applications of CNNs revolve around detection or image enhancement to aid in wildlife exploration \cite{han2020marine,sun2017underwater,wang2019uwgan}. Specifically for underwater docking, the list is reduced \cite{liu2018detection,yahya2017detection} since supervised learning requires large amounts of data to train on, which is sparse for the underwater domain. This poses a challenge for developing supervised machine-based underwater vision algorithms \cite{wang2019uwgan}. One way to resolve this challenge is by creating artificial datasets, as shown in \cite{chavez2022cnn,chavez2022efficient,alvarez2019generation}, where underwater images are created artificially using style transfer techniques.

Another challenge associated with the marine environment is the cost and risks during field experiments. The underwater environment in which AUV operates is considered dangerous, not only because of the currents and/or obstacles that a vehicle might encounter but also because of the water clarity conditions. This is especially a problem in coastal regions or areas with high human activity \cite{abrosimov2022machine}. Underwater docking can be a challenging and potentially dangerous operation due to factors such as water currents, visibility, and the need for precise positioning; thus, having a simulation tool capable of simulating different environmental and water conditions is valuable since it can help to identify potential issues and provide insight into the effectiveness of different methods before testing in the field. There are several physics engines and simulation environments available for the simulation of robots, such as MuJoCo\cite{todorov2012mujoco}, ODE \cite{smith2007open}, Gazebo, and Unity. These environments not only provide a way of simulating the motion of the vehicle and contact forces but also provide the physics associated with the sensors used for the control of the robot. All these tools provide a reliable way to test ground or even aerial vehicles; however, the underwater domain requires the simulation of sensors that are not commonly used in other domains, such as sonars or acoustic modems. In addition, sensors such as cameras work differently underwater, mainly due to the way light propagates underwater \cite{liu2018detection}. This is critical to have a realistic underwater docking scenario. The underwater domain simulation environments such as the ones developed in \cite{bhat2019towards,cieslak2019stonefish,Manhaes_2016,kermorgant2014dynamic} provide ways to simulate the dynamics of diverse underwater vehicles; however, they cannot model the scattering and absorption effects underwater effectively. 

Another difficulty in deploying deep learning algorithms is the required computational resources.
Although some deep learning algorithms are known to be efficient for image processing tasks, they still require many parameters to perform well, implying that an embedded device used for implementation requires enough memory to allocate the network parameters \cite{gou2021knowledge}.
As a result of this issue, model compression techniques have been developed to achieve comparable performance with fewer resources \cite{chen2017learning,jafari2021annealing}.
The teacher-student architecture, which requires two networks, is an example of compression.
The teacher network is typically deep because large networks have lower generalization errors \cite{nakkiran2021deep}, whereas the student network is shallower and is trained to mimic the teacher behavior \cite{gou2021knowledge}. 

This paper presents an efficient underwater docking localization system validated in a neural network based realistic simulation environment. Different architectures are trained and evaluated in terms of their localization and classification errors, as well as their inference time. The best architecture is selected to train a deep model, which will serve as a reference (teacher) for a shallower network (student) so that the knowledge from the deeper model is distilled into the shallower network. This has the advantage of improving the performance of the student network without the need for a large number of parameters. After that, we evaluate the impact of learning-based detection against classical vision methods for underwater docking. Finally, we discuss some of the results obtained after deploying the selected method into the Iver3 AUV. 

The remainder of this work covers the proposed underwater docking detection architectures evaluated in Sec.~\ref{sec:uw_dock}. The simulation environment and the process to generate realistic underwater-looking images are described in Sec.~\ref{sec:sim}. The results are discussed in Sec.~\ref{sec:results} followed by a conclusion in Sec.~\ref{sec:conc}.

\section{Underwater Docking}\label{sec:uw_dock}
\subsection{Docking Strategy}
\begin{figure*}
\centering
    \includegraphics[width=1.0\textwidth]{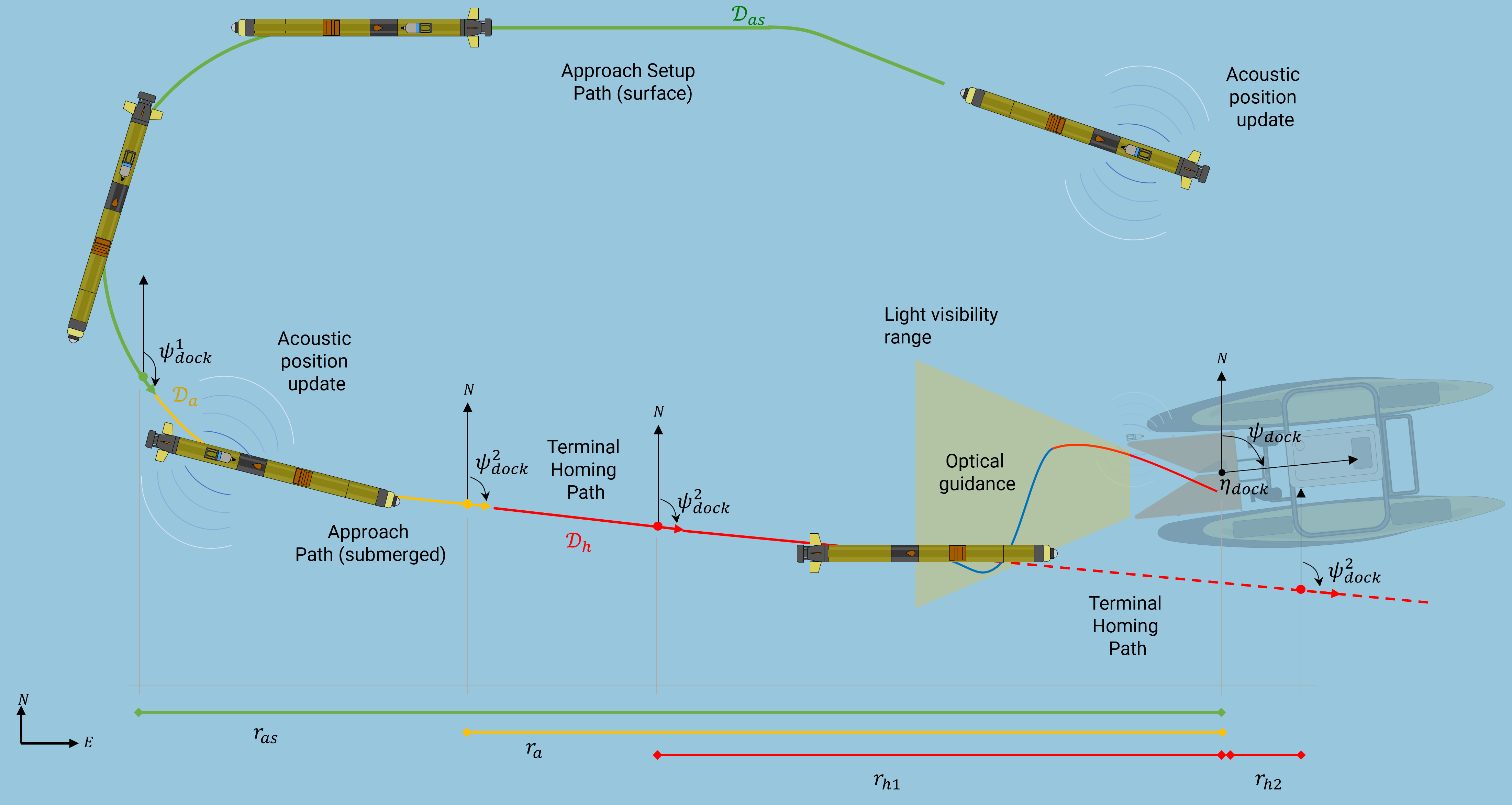}
\caption{Underwater docking strategy composed of several paths that guide the AUV towards the ASV carrying the docking station. Once the vehicle is close enough and the light beacon is in a visible range, the AUV is optically guided. If at any point the light is out of the frame, then the AUV switches back to follow the previously generated path.}
\label{fig:docking_strategy}
\end{figure*}

The docking strategy presented in this paper involves multiple stages, as mentioned in Sec.~\ref{sec:intro}. The main idea behind this is to increase the probability of success by aligning the vehicle to the docking station, reducing the complexity of the docking maneuver during the terminal homing. In this work, the docking station is carried by an Autonomous Surface Vehicle (ASV); once the vehicle finds a suitable location to wait for the AUV, it starts a maneuver known as station keeping, in which the ASV will keep its position and heading creating ideal conditions for the vehicle to dock as shown in \cite{chavez2023asv}. The ASV can be either configured to broadcast its position continuously or to send it upon request. In any case, the AUV will wait for this message and will trigger the start of the docking maneuver after receiving the ASV's position and orientation. During the docking maneuver, a Dubins path planner and an Integral Line of Sight (ILOS) controller as the one in \cite{page2021underwater} will control the AUV trajectory. During the terminal homing, the AUV will receive an additional control signal from the neural network-based light follower; this is illustrated in Fig.~\ref{fig:docking_strategy}. The first stage of the docking maneuver, which is known as approach setup, starts with the generation of a Dubins path $\mathcal{D}_{as}$ with a final waypoint $r_{as}>0$ meters behind $\pmb{\eta}_{dock}$, and the same orientation as the docking station. This stage is completed on the surface to get a GPS fix before submerging. Once the approach setup stage is completed, the approach stage starts; in this stage, a path $\mathcal{D}_{a}$ is generated. This path has a final waypoint $r_{a}<r_{as}$ meters before $\pmb{\eta}_{dock}$ at the same depth $z_{dock}$ and orientation of the docking station. The terminal homing stage starts right after the approach path has been completed. This segment of the docking maneuver consists of two waypoints: one $r_{h1}$ meters before $\pmb{\eta}_{dock}$ and one $r_{h2}$ meters after; this forces the generation of a linear path that in the best scenario will be aligned to the centerline of the docking station. At any point during the maneuver, if a new acoustic position message is available, the path generation of the subsequent stage will consider the new value in an attempt to keep up with the slowly moving ASV. It is important to consider sensor errors, which implies that the path $\mathcal{D}_{h}$ will slightly deviate from the docking station centerline. This is why the docking detection in the last part of the docking maneuver becomes crucial since it helps to correct the localization error, driving the vehicle toward the docking envelope. If the docking maneuver fails, the terminal homing will be completed, and then the docking maneuver will be started from the approach stage. The overall docking strategy, the different paths $\mathcal{P}=\{\mathcal{D}_{as},\mathcal{D}_{a},\mathcal{D}_{h}\}$, and the docking maneuver are shown in Fig.~\ref{fig:docking_strategy}. 

\subsection{Detection Architecture}
\begin{figure}
    \centering
     \includegraphics[width=\columnwidth]{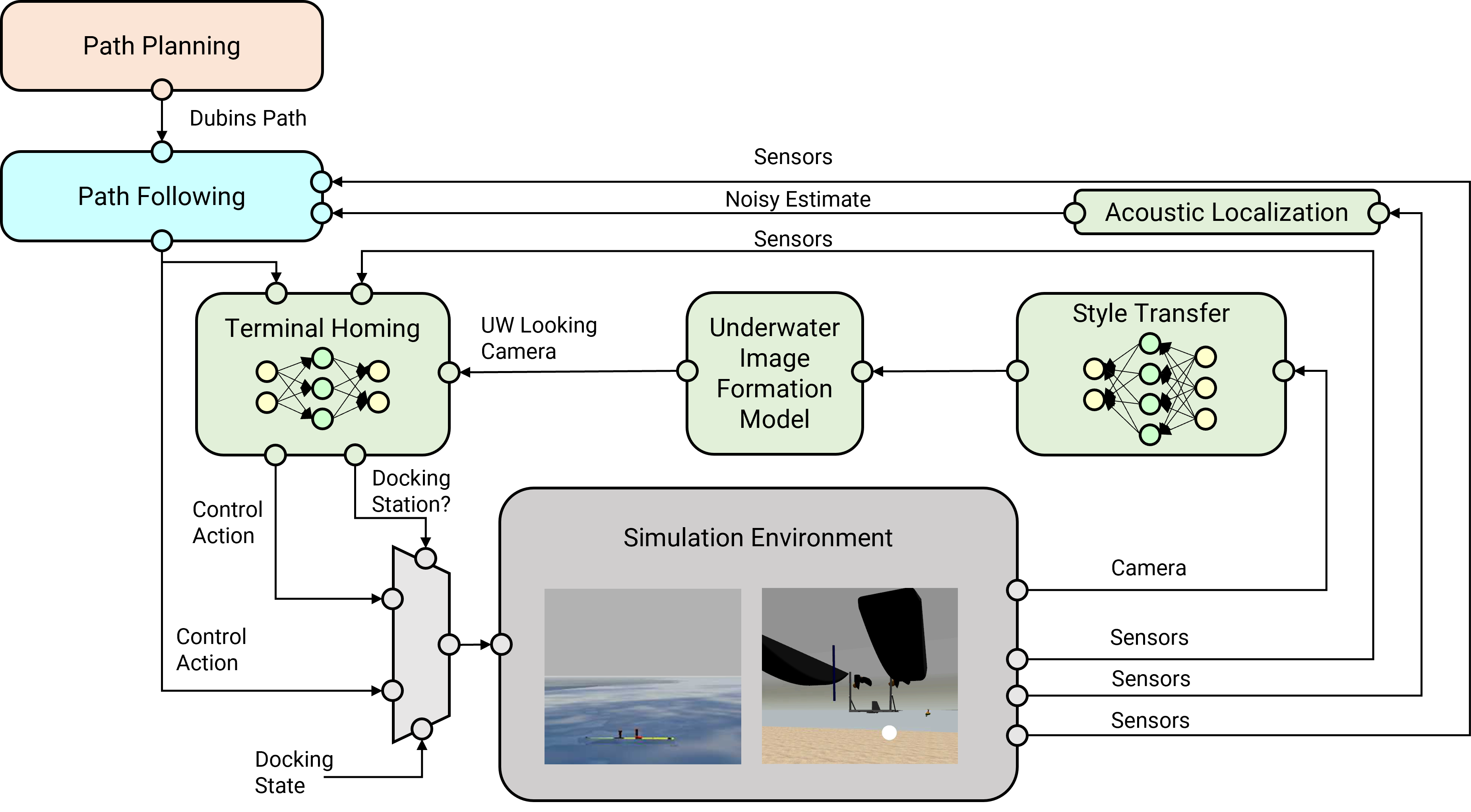}
     \caption{Architecture of the docking system. During approach setup and approach stages, path planning and path follower components generate a path to align the vehicle towards the docking station using the information received through the acoustic modem. The AUV is equipped with a camera. The output of the camera is processed to generate an underwater-looking image fed into the underwater docking detection algorithm. This procedure is employed during the terminal homing stage to perform a fine adjustment of the vehicle pose to dock successfully. If a failure is detected the procedure is repeated in the missed approach stage.}
     \label{fig:uwd_sys_arch}
\end{figure}

After reaching the last stage of the docking procedure, docking detection becomes essential, since it will reduce the localization error, and adapt to ASV's slight changes online at the required frequency. The proposed underwater docking detection architecture shown in Fig.~\ref{fig:uwd_sys_arch} consists of an AUV equipped with an acoustic modem and a camera. The acoustic modem gets an initial (but low-rate) estimate of the docking station pose. The AUV uses this information to calculate a Dubins path towards the mobile docking station. Once within range, the path following strategy switches to visual guidance using the camera, which identifies the docking position in the image frame with the help of a light beacon that is mounted to the docking station. The visual guidance serves to fine tune (at a higher rate) the pose of the AUV. In this work, the underwater docking detection is done using a neural network based model that predicts if the docking station is present and estimates the position of the docking station in the image frame. This, together with the acoustic guidance is used to compute a direction for the AUV to follow during the terminal homing stage (last $~15$ m). If, during the terminal homing, the docking station is not visible, the navigation is guided purely based on acoustic guidance, providing a safety mechanism for the procedure. The procedure finishes when the latching is achieved. If a failure is detected, the vehicle retries until it finally latches onto the dock. 


\subsection{Dataset Generation}
\begin{figure}
\centering
\begin{subfigure}{0.45\columnwidth}
    \includegraphics[width=\textwidth]{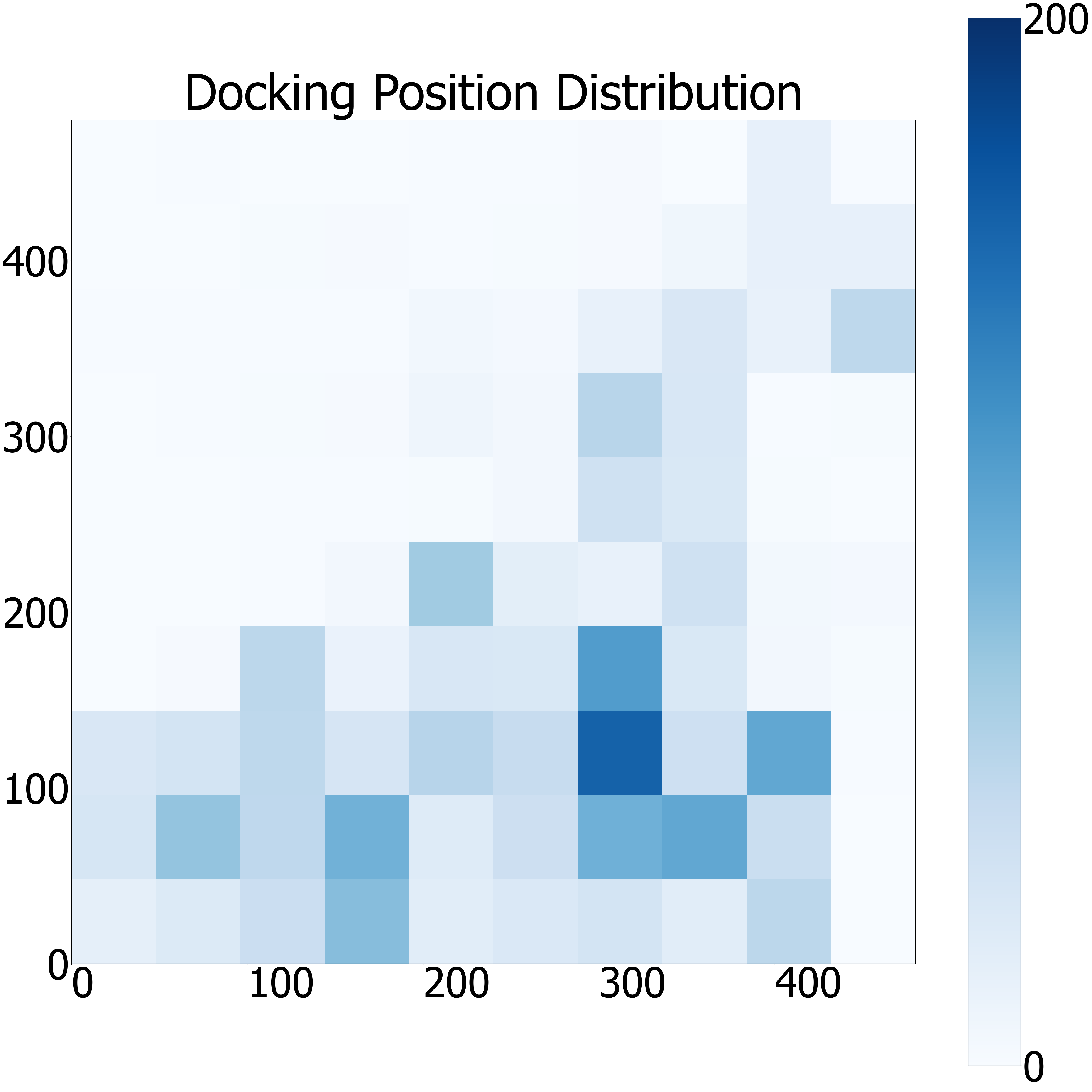}
    \caption{}
    \label{fig:dataset_dist_original}
\end{subfigure}
\hfill
\begin{subfigure}{0.45\columnwidth}
    \includegraphics[width=\textwidth]{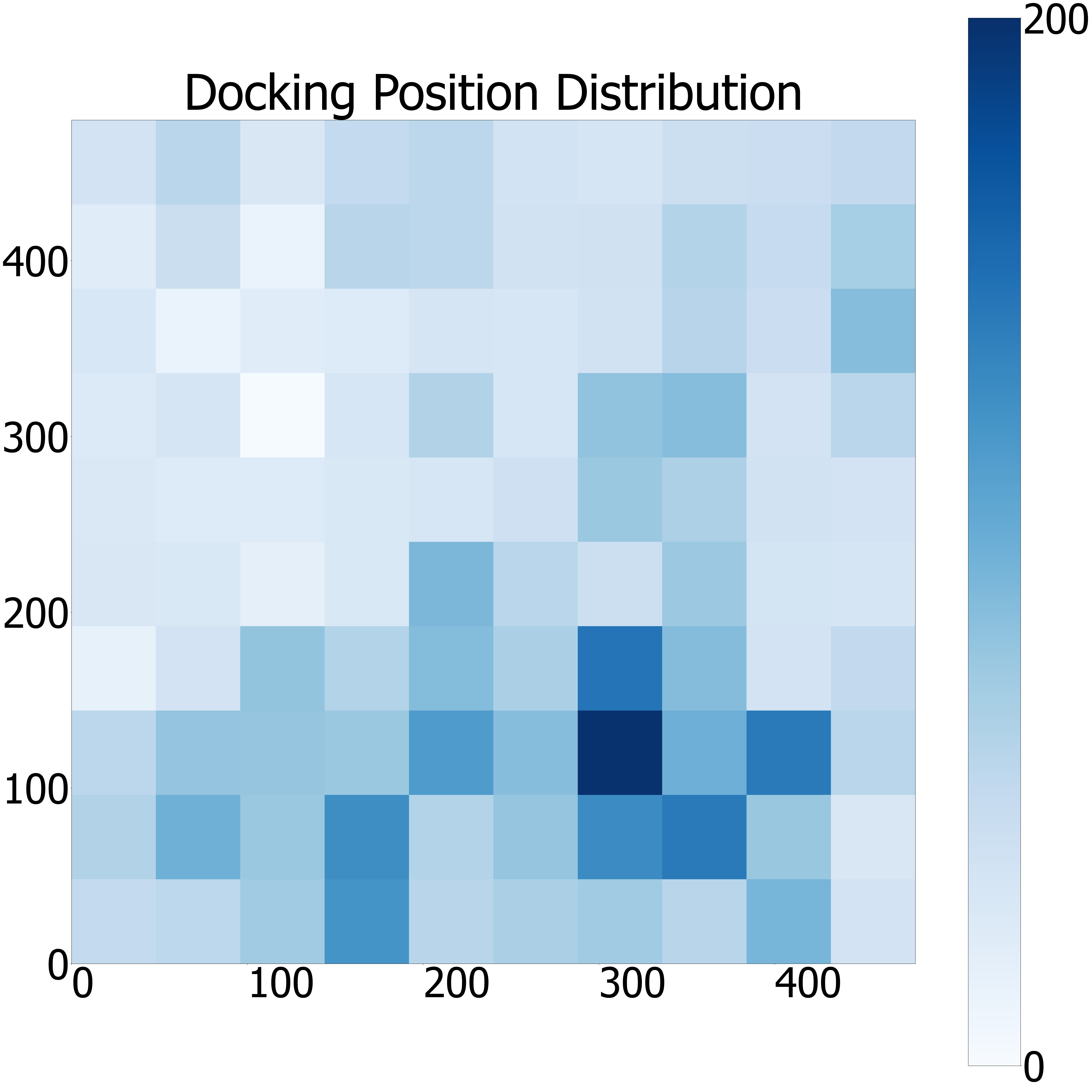}
    \caption{}
    \label{fig:dataset_dist_corrected}
\end{subfigure}      
\caption{Distribution of the docking position in the image frame on the original dataset (a), and the final dataset (b) after the generation of artificial images. As can be observed, the distribution of the docking position in (b) is more evenly distributed to avoid any bias in the detection.}
\label{fig:dataset_dist}
\end{figure}

The dataset used to train the underwater docking detection algorithm comprises real and artificial images $X=\{X^{real}, X^{ast}, X^{cgan}\}$. The artificial images are generated using either artistic style transfer \cite{gatys2015neural}, or Cycle GAN \cite{zhu2017unpaired} images as described in \cite{chavez2022cnn,chavez2022efficient}. The portion of the dataset corresponding to real images $|X^{real}| = 4828$ is equally distributed across the two classes, docking and no docking. The distribution of the position of the docking station in the image frame is shown in Fig.~\ref{fig:dataset_dist_original}; such a distribution suggests that the training could lead to a bias in the detection model. To mitigate this effect, $7000$ artificial images with a light position evenly distributed across the image frame are generated, resulting in a more balanced dataset across the two classes and throughout the image frame. The distribution of light positions, including the artificial images $|X^{cgan}| + |X^{ast}| = 7000$ is shown in Fig.~\ref{fig:dataset_dist_corrected}. After increasing the size of the dataset, we create a threefold split consisting of $70$\% for training, $20$\% for validation, and $10$\% for testing. During training and validation, the dataset is also augmented in 10 ways to allow translation, rotation, and color invariance for the detection. To assess the generalizability of the network, and the efficacy of the artificial data, the testing data is purely composed of real data $X^{test} \subset X^{real}$.

\subsection{Architecture Comparison}
In order to conduct underwater docking detection, this study compares various neural network architectures in terms of accuracy, localization error, number of parameters, memory usage, and inference time. In particular, CNN, Pyramidal CNN, Residual Blocks, and Transformers designs are considered. This comparison aims to determine the best architecture for the underwater docking task, considering that the detection is part of the control loop of the vehicle. This implies that a relatively fast and lightweight algorithm is desirable. Each architecture being compared has advantages and disadvantages, so it is essential to compare several aspects to determine the most suitable architecture for the underwater docking detection task. In this work, we consider detection loss, classification loss, memory usage, and inference time to assess the most suitable architecture.

\subsubsection{CNN}
CNNs are a type of neural network architecture that has gained attention in computer vision since they can automatically learn spatial features from images. They leverage three basic principles: sparse interactions, parameter sharing, and equivariant representations \cite{goodfellow2016deep}. CNNs consist of several convolutional layers that apply filters to the input image, extracting local features. The output is then fed into a nonlinear activation function and then to a pooling layer, which reduces the spatial resolution of the features. After being applied several times, this process results in a hierarchical representation of the input image, which can finally be fed into a Fully Connected Network (FCN) to produce the desired output.

\subsubsection{Residual Network}
Residual networks, also known as ResNets are an architectural type of neural networks widely used in the field of computer vision. The key idea behind ResNets is the introduction of skip connections, which allow data to bypass layers and flow directly to the output in the network \cite{he2016deep}. This can alleviate the vanishing gradient problem, which can negatively impact the training of deep neural networks. The skip connections enable the network to learn residual functions, capturing the difference between the input and the desired output. ResNets leverage residual learning to build deep networks, enabling improved performance and faster convergence. The reformulation presented by residual learning allows an identity mapping, if optimal, in such a way that the optimizer drives the weights of the multiple nonlinear layers toward zero to approach identity mappings. This ability allows enhanced accuracy and the ability to handle complex tasks in computer vision.

\subsubsection{Pyramidal Convolution}
Nearly every recent state-of-the-art architecture for visual recognition is based on CNNs \cite{goodfellow2016deep,duta2020pyramidal}. Most CNNs use a small kernel size, generally driven by the computational cost of increasing this. To overcome the problem, CNNs use several layers with a pooling layer to downsample the feature maps, and thus, gradually increase the receptive field of the network. According to \cite{zhou2014object}, although this is true theoretically, it can be shown empirically that the actual receptive field is, in some cases, more than 2.7 times smaller. Another problem arises while downsampling the input at every layer since useful details are lost due to the small receptive field before downsampling. To address these problems \cite{duta2020pyramidal} proposes a pyramidal convolution (PyConv), which contains different kernels with varying sizes and depths. This has, as a result, an increased receptive field. The PyConv in \cite{duta2020pyramidal} is shown to preserve a similar number of parameters as a standard convolution. PyConv utilizes a 1×1 2D convolution to downsize the input feature maps from $FM_e$ to $FM_i$. Subsequently, the reduced feature maps $FM_i$ are passed through $n$ separate convolution operations with varying kernel sizes $K_1,...,K_n$. Each of these convolution operations yields $FM_{oi}$ feature maps, which together sum up to $FM_{o}$. To maintain computational efficiency, grouped convolution is employed for each kernel. Larger kernels employ more groups during convolution, thereby reducing the connection between input and output, resulting in fewer computational operations. Following this, a 1×1 2D convolution is applied to restore the initial number of feature maps $FM_e$. Each 2D convolution block is followed by batch normalization and a ReLU activation function. Lastly, a shortcut connection, similar to those found in residual blocks, is included. In our work, we implement two versions of the PyConv block, one with a residual connection and one without.

\subsubsection{Transformer}

Transformers have recently revolutionized natural language processing (NLP) \cite{vaswani2017attention}, and more recently, various other tasks such as computer vision \cite{dosovitskiy2020image}. Unlike traditional recurrent or convolutional neural networks, transformers rely on a self-attention mechanism to process sequential data. The key idea behind transformers is the concept of attention, which allows the model to focus on different parts of the input sequence when making predictions. The transformer architecture consists of an encoder and a decoder, both composed of multiple layers of self-attention and feed-forward neural networks. During training, transformers learn to attend to relevant information and capture long-range dependencies by assigning different weights to different parts of the input sequence. This self-attention mechanism enables transformers to model context more effectively and handle variable-length input sequences. With their ability to capture global dependencies and effectively process sequential data, transformers have become the de-facto architecture for many NLP tasks. Inspired by the successes of transformers in NLP, \cite{dosovitskiy2020image} experimented with applying a standard
transformer to images. To do it, the image is split into patches and fed to the transformer as a sequence of linear embeddings of these patches. Image patches are treated as tokens (words) in an NLP application. Ultimately, the model is trained to do image classification in a supervised fashion.

\subsection{Knowledge Distillation}
Deeper networks tend to generalize better if correctly trained because they have a bigger capacity \cite{nakkiran2021deep}, but this comes at the expense of longer inference times and higher memory utilization. However, for minor jobs, that level of complexity may not be required; consequently, model compression can be used to reduce inference times and memory usage. Some research in the field of knowledge distillation, a sort of model compression, has demonstrated that a shallow model trained to mimic the behavior of a larger network can recover some, if not all, of the accuracy drop \cite{chen2017learning,jafari2021annealing}.

This work proposes the usage of a teacher-student knowledge distillation as a method to reduce the complexity and the number of parameters of the final architecture. This will ease the network deployment on a more constrained embedded device. There are several methods of knowledge distillation; in the area of computer vision, most of them focus on the classification task rather than a regression task (object localization) \cite{gou2021knowledge}. The implementation of this work utilizes a loss function $\mathcal{L}(\theta)$ for the student network upper bounded by the teacher, instead of learning it directly from the labeled data as in \cite{chen2017learning}. 

\begin{equation}
    \label{eq:kd_loss}
    \mathcal{L}(\theta) = \frac{1}{B}\{\sum_{i=1}^{B} (\|Y_i-f(x_i;\theta)\| + v\mathcal{L}_T)\}
\end{equation}

where $Y_i$ is the ground truth, $\mathcal{L}_T=\|Y_i-f(x_i;\theta)\|$ if $\|Y_i-f(x_i;\theta)\| > \|Y_i-f(x_i;\theta_T)\|$, and $\mathcal{L}_T=0$ otherwise. The parameter $v$ is used to adjust the weight of $\mathcal{L}_T$ on the final loss.

\section{Underwater Simulation}\label{sec:sim}
\subsection{AUV Model}
To understand the dynamics of the AUV a 6-DOF mathematical model is used to describe the surge, sway, heave, roll, pitch, and yaw motion. This model assumes that the vehicle is submerged the whole time, and thus the center of gravity and the center of buoyancy coincide. It is also assumed that the vehicle is perfectly trimmed such that the density of the vehicle and the density of the surrounding fluid are the same. The vehicle position $\pmb{\eta}=[\pmb{\eta}_1^T,\pmb{\eta}_2^T]^{T}$ can be described in terms of the earth-fixed frame coordinates $\pmb{\eta}_1=[x,y,z]^{T}$, and the Euler angle vector $\pmb{\eta}_2=[\phi,\theta,\psi]^{T}$. The velocity $\pmb{V}=[\pmb{V}_1^T,\pmb{V}_2^T]^{T}$ can be described in terms of the body-fixed linear velocity $\pmb{V}_1=[u,v,w]^{T}$, and the body-fixed angular speed $\pmb{V}_2=[p,q,r]^{T}$. The fin-related hydrodynamic contribution $\pmb{\tau}_{hydr}=[\pmb{\tau}_1^T,\pmb{\tau}_2^T]^T$ is represented in terms of the forces $\pmb{\tau}_1=[\tau_x,\tau_y,\tau_z]^T$ and moments $\pmb{\tau}_2=[\tau_\phi,\tau_\theta,\tau_\psi]^T$. Finally, the contribution of the thruster is included in the model as $\pmb{\tau}_{thrust}=[f_{thrust},0,0,0,0,0]^T$. The dynamics model of the Iver3 can then be described as:

\begin{align}
    \dot{\pmb{\eta}}=\pmb{J}(\pmb{\eta}_2)\pmb{V}\\
    \pmb{M}\dot{\pmb{V}} + \pmb{C}(\pmb{V})\pmb{V} + \pmb{D}(\pmb{V})\pmb{V} = \pmb{\tau}_{hydr} + \pmb{\tau}_{thrust} \label{eq:dyn_model}
\end{align}

where $\pmb{J}(\pmb{\eta}_2)$ is the rotation matrix from the body-fixed frame to the earth-fixed frame. 

\begin{align}
    \pmb{J}(\pmb{\eta}_2) = \begin{bmatrix}
    \pmb{J}_1(\pmb{\eta}_2) & \pmb{0}_{3\times 3}\\
    \pmb{0}_{3\times 3} & \pmb{J}_2(\pmb{\eta}_2)
    \end{bmatrix}
\end{align}

with $\pmb{J}_1(\pmb{\eta}_2)$ and $\pmb{J}_2(\pmb{\eta}_2)$ defined as:

\begin{align}
    \pmb{J}_1(\pmb{\eta}_2) =
    \begin{bmatrix}
    \cos{\psi} \cos{\theta} & -\sin{\psi}\cos{\phi} + \cos{\psi}\sin{\theta}\sin{\phi} &  \sin{\psi}\sin{\phi} + \cos{\psi}\cos{\phi}\sin{\theta} \\
    \sin{\psi} \cos{\theta} &  \cos{\psi}\cos{\phi} + \sin{\phi}\sin{\theta}\sin{\psi} & -\cos{\psi}\sin{\phi} + \sin{\theta}\sin{\psi}\cos{\phi} \\
    -\sin{\theta}           &  \cos{\theta}\sin{\phi}                                  & \cos{\theta}\cos{\phi}
    \end{bmatrix}\\
    \pmb{J}_2(\pmb{\eta}_2) =
    \begin{bmatrix}
    1 & \sin{\phi}\tan{\theta}  & \cos{\phi}\tan{\theta}\\
    0 & \cos{\phi}              & -\sin{\phi}\\
    0 & \sin{\phi}/\cos{\theta} & \cos{\phi}/\cos{\theta}\\
    \end{bmatrix}
\end{align}

where $\sin{\cdot}=sin(\cdot),\cos{\cdot}=cos(\cdot)$ and $\tan{\cdot}=tan(\cdot)$. In Eq.~\ref{eq:dyn_model} the mass $\pmb{M}$ and the Coriollis $\pmb{C}(\pmb{V})$ matrices can be expressed in terms of the rigid-body and added mass components $\pmb{M}=\pmb{M}_{RB}+\pmb{M}_{A}$ and $\pmb{C}(\pmb{V})=\pmb{C}_{RB}(\pmb{V})+\pmb{C}_{A}(\pmb{V})$. The remaining part of Eq.~\ref{eq:dyn_model} $\pmb{D}$, which is the damping matrix, can be reduced to its linear components for speeds below $2$~m/s as in \cite{wang2015dynamic}. Therefore, the damping matrix can be defined as $\pmb{D}=-diag\{X_u,Y_v,Z_w,K_p,M_q,N_r\}$.

\vspace{0.3cm}
\subsection{Simulation Setup}
\begin{figure}[t]
\footnotesize
\centering
\includegraphics[width=0.42\textwidth]{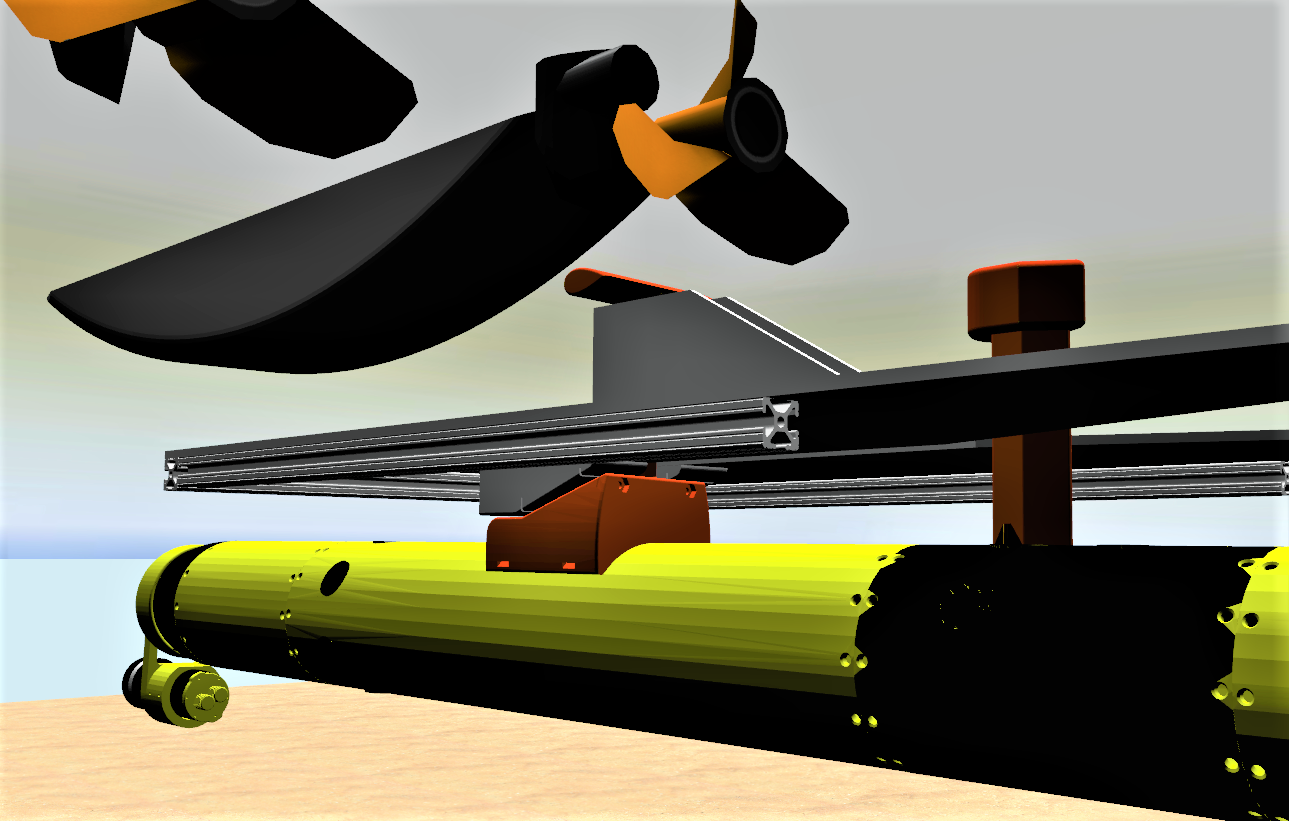}
\caption{Simulated docking environment.}
\label{fig:sim_gazebo}
\end{figure}

The simulation used to test the overall underwater docking strategy was implemented on Gazebo and the Robot Operating System (ROS). The ASV, the docking station, and the AUV are modeled in SOLIDWORKS and Blender. In addition, plugins from the UUV simulator \cite{Manhaes_2016} and VRX simulator \cite{bingham19toward} are used to include the hydrodynamic-related components in the motion of the vehicles. The docking station and light beacon are attached to the ASV. The diagram in Fig.~\ref{fig:sim_gazebo} shows the AUV, the docking station and the ASV in a successful docking attempt. In this architecture, Gazebo and ODE are in charge of the simulation of the rigid-body dynamics, while the control strategy is implemented entirely on ROS as can be seen in Fig.~\ref{fig:uwd_sys_arch}. To close the gap between the simulation and field experiments, the acoustic localization message transmitted from the ASV to the AUV was simulated through a ROS node. This component included two important aspects observed in the field experiments; one is the GPS and compass error, and the other is the probability of success in the transmission/reception of the message. To account for these phenomena, the measured dock position includes an error term $\pmb{\epsilon}\sim \mathcal{N}(0,\pmb{\sigma}^2_{dock})$ simulating error in the GPS and compass. Additionally, to simulate corrupted communication resulting in a faulty transmission, a Bernoulli distribution determines if the acoustic message is correctly received with a success probability of $p$ and an error probability of $q=1-p$. The acoustic localization node is configured to run at $0.33$Hz, which emulates the observed speed between the pair of acoustic modems SeaTrac X150 and X110.


\subsection{Underwater Imagery}
The Gazebo simulator provides the required plugins to simulate a camera. This plugin allows modifying several camera parameters, such as width, height, field of view (FOV), or noise. This can endow the simulation with a more realistic performance. However, images underwater suffer from attenuation and absorption effects, resulting in inaccurate colors and low contrast \cite{berman2020underwater}, which is not simulated in the default plugin. To generate visually immersive underwater scenes and to provide a realistic test environment, we trained an image-to-image translation network based on Pix2Pix \cite{isola2017image}, using the cycle consistency from Cycle GAN \cite{zhu2017unpaired}. To obtain the paired dataset to train Pix2Pix, it was necessary to collect a dataset from the simulation environment $X^{sim}$ and then find the closest match from $X^{real}$. Once the paired dataset is available, the network was trained for $500$ epochs to find the mappings $G$ and $F$, such that $G(x^{sim})\approx x^{real}$ and $F(G(x^{sim})))\approx x^{sim}$. The model learned from the Pix2Pix, although it is good to transfer the underwater style onto the Gazebo simulated image, it cannot simulate the attenuation. To bring that into the simulation, the underwater image formation model described in \cite{akkaynak2018revised} was used; this also gives the ability to simulate different attenuation factors according to the Jerlov water types described \cite{jerlov1977classification}. The obtained model described by the following equation is integrated into the simulation, as shown in Fig.~\ref{fig:uwd_sys_arch}.

\begin{align}
     x_{uw} &= K_{gen}G(x_{sim})+K_{sim}x_{sim}\\
     x_{att} &= x_{uw}\cdot e{^{-\beta_c\cdot d}} + \beta_c^{\infty}\cdot(1-e{^{-\beta_c\cdot d}})
\end{align}

where $K_{gen}$ and $K_{style}$ are positive weights such that $K_{gen} + K_{style} = 1$ in charge of controlling the influence of $x_{sim}$ and $G(x_sim)$ on the unattenuated output image $x_{uw}$. Then the attenuated image $x_{att}$ is generated through $x_{uw}$ and the value of backscatter at infinity, also termed veiling light $\beta_c^{\infty}$. The parameters $\beta_c$, also known as attenuation coefficients, and the range along the line of sight $d$ are used to combine $\beta_c^{\infty}$ and $x_{uw}$.

\section{Results}\label{sec:results}

\subsection{Undewater Docking Detection}
\subsubsection{Dataset Generation}
\begin{figure}[!hb]
\centering
\begin{subfigure}{0.3\columnwidth}
    \includegraphics[width=\textwidth]{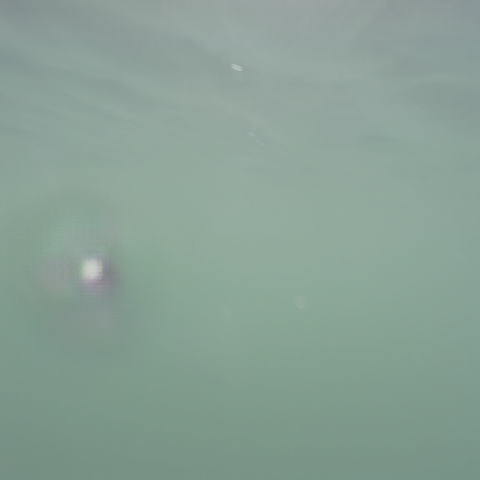}
    \caption{}
    \label{fig:dataset_examples_cgan_1}
\end{subfigure}
\begin{subfigure}{0.3\columnwidth}
    \includegraphics[width=\textwidth]{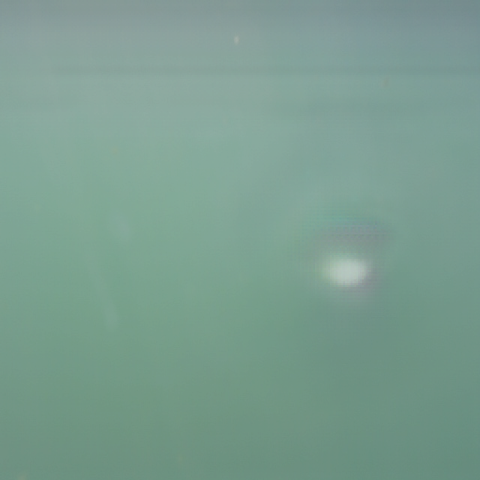}
    \caption{}
    \label{fig:dataset_examples_cgan_2}
\end{subfigure}\\

\begin{subfigure}{0.3\columnwidth}
    \includegraphics[width=\textwidth]{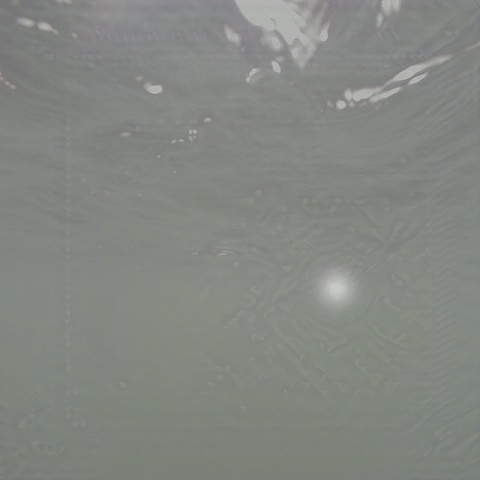}
    \caption{}
    \label{fig:dataset_examples_ast_1}
\end{subfigure}
\begin{subfigure}{0.3\columnwidth}
    \includegraphics[width=\textwidth]{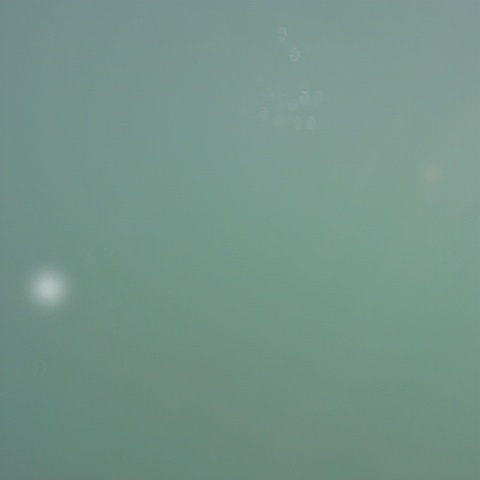}
    \caption{}
    \label{fig:dataset_examples_ast_2}
\end{subfigure}
\caption{Examples of images generated to train the different networks. (a) and (b) are images generated using the Cycle GAN network, while (c) and (d) are images generated using artistic style transfer.}
\label{fig:dataset_examples}
\end{figure}

The dataset used in this paper comes from real experiments but also from artificial data generated using Cycle GAN and artistic style transfer according to the process described in \cite{chavez2022cnn,chavez2022efficient}. The artificial data was generated to augment the dataset in such a way that the position of the docking station was distributed more evenly across the image frame. Figure~\ref{fig:dataset_examples} provides a visual representation of the artificial image generation process using both the artistic style transfer network and the Cycle GAN network. The final dataset denoted as $X=\{X^{cgan},X^{ast},X^{real}\}$ consisted of $|X| = 11828$ images, with  $|X^{cgan}| + |X^{ast}| = 7000$ and $|X^{real}| = 4828$. The dataset $X$ was then split such that $|X^{train}|=8279$, $|X^{eval}|=2366$ and $X^{test}=|1183|$. These subsets represent 70\%, 20\%, and 10\% of the total dataset respectively. To assess the ability of the networks to generalize on real data $X^{test}$ is puerely composed of real data, while $X^{train}$ and  $X^{test}$ are composed of $34\%$ real and $66\%$ artificial data.

\subsubsection{Architecture Comparison}
\begin{figure}[t]
\centering
\begin{subfigure}{0.48\columnwidth}
    \includegraphics[width=\textwidth]{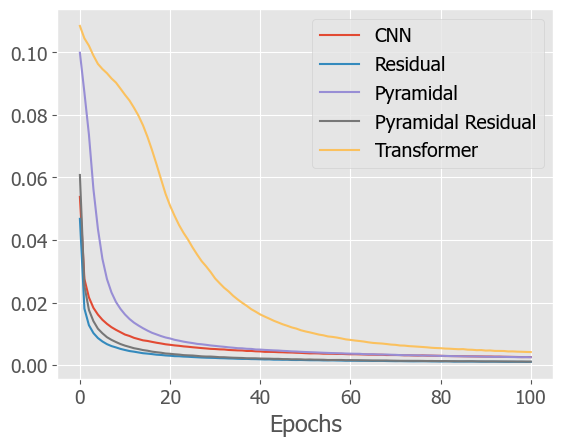}
    \caption{}
    \label{fig:train_arch_l1}
\end{subfigure}
\hfill
\begin{subfigure}{0.48\columnwidth}
    \includegraphics[width=\textwidth]{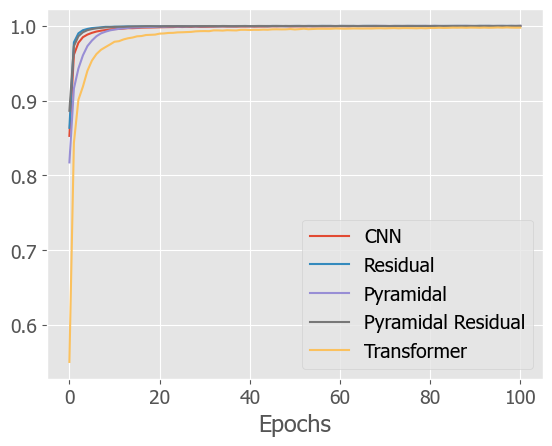}
    \caption{}
    \label{fig:train_arch_acc}
\end{subfigure}      
\caption{Training L1 loss (a) and accuracy (b) of the underwater docking detection networks evaluated.}
\label{fig:train_arch_loss}
\end{figure}

\begin{figure}[t]
\centering
\begin{subfigure}{0.49\columnwidth}
    \includegraphics[width=\textwidth]{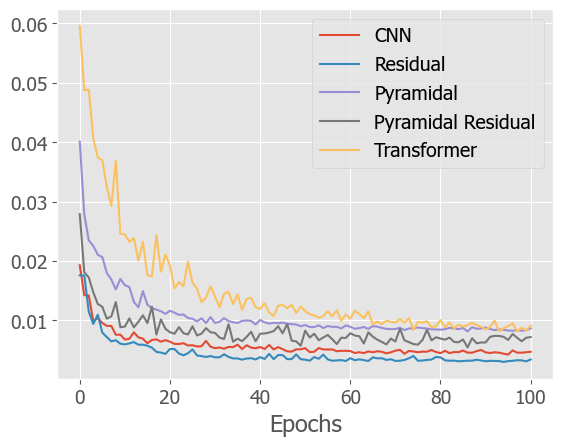}
    \caption{}
    \label{fig:train_arch_l1}
\end{subfigure}
\hfill
\begin{subfigure}{0.49\columnwidth}
    \includegraphics[width=\textwidth]{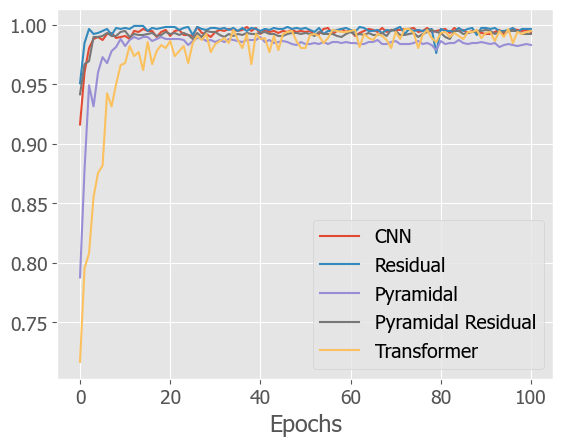}
    \caption{}
    \label{fig:train_arch_acc}
\end{subfigure}      
\caption{Testing L1 loss (a) and accuracy (b) of the underwater docking detection networks evaluated.}
\label{fig:eval_arch_loss}
\end{figure}

Five different architectures were trained for $100$ epochs on the dataset $X^{train}\subset X$ composed of real and artificial data to determine the most suitable architecture for underwater docking detection. The learning rate used for all the networks was $4e-6$, and a batch size of 8. The metrics considered to evaluate the different architectures were L1 loss, accuracy, inference time, and number of parameters. Fig.~\ref{fig:train_arch_loss} and Fig.~\ref{fig:eval_arch_loss} illustrate the loss of the different architectures on the train $X^{train}$ and test dataset $X^{test}\subset X^{real}$ respectively, which is only composed of real data. This validates the generalizability of the architectures.
The transformer network exhibits good accuracy but has the poorest performance in localization. This can be related to the $16\times16$ embedding patches used to decompose the image into smaller portions of information. Additionally, it has much less image-specific inductive bias than other architectures such as CNN or Residual. Unlike these networks, the transformer needs to learn the spatial relation between the patches from scratch, which is observed in the slow learning rate in the early stages of the training in Fig.~\ref{fig:train_arch_loss}. Although the final L1 loss error is under 10\% of the image width for all the networks, the pyramidal shows a slightly higher error than CNN or residual networks. We attribute the reduced performance on the pyramidal network to the initial $7\times7$ convolution layer with a stride of $2$, which reduces the spatial resolution and, thus, the granularity of docking detection. After adding the residual shortcut to the pyramidal network, we can observe an improvement in the performance, compared to the pyramidal network, mainly attributed to the possibility of skipping layers when needed. This effect is also observable in the CNN and residual networks, which have the best performance overall. After adding the identity shortcut to a conventional CNN block, the network's performance is boosted. This can be seen in both Fig.~\ref{fig:train_arch_loss} and Fig~\ref{fig:eval_arch_loss}, where the residual network was the best performance overall.

\begin{figure}[H]
\centering
\begin{subfigure}{0.42\columnwidth}
    \includegraphics[width=\textwidth]{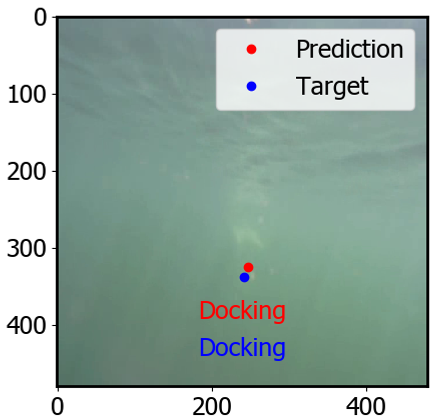}
    \caption{}
    \label{fig:qual_d_cnn}
\end{subfigure}
\hfill
\begin{subfigure}{0.42\columnwidth}
    \includegraphics[width=\textwidth]{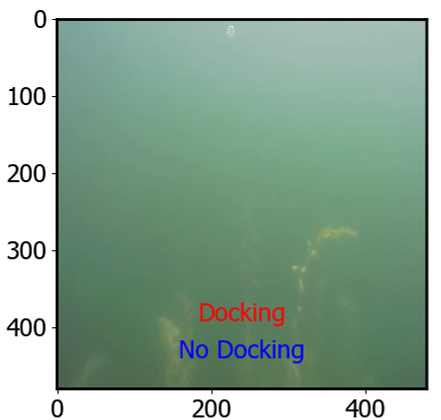}
    \caption{}
    \label{fig:qual_nd_cnn}
\end{subfigure}
\begin{subfigure}{0.42\columnwidth}
    \includegraphics[width=\textwidth]{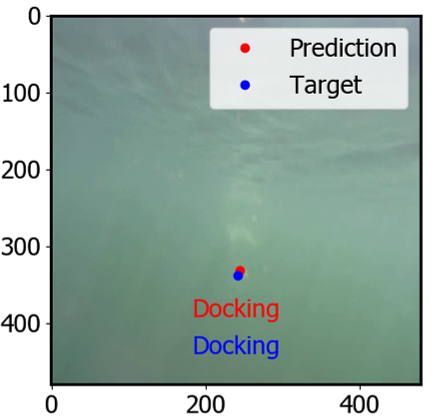}
    \caption{}
    \label{fig:qual_d_res}
\end{subfigure}
\hfill
\begin{subfigure}{0.42\columnwidth}
    \includegraphics[width=\textwidth]{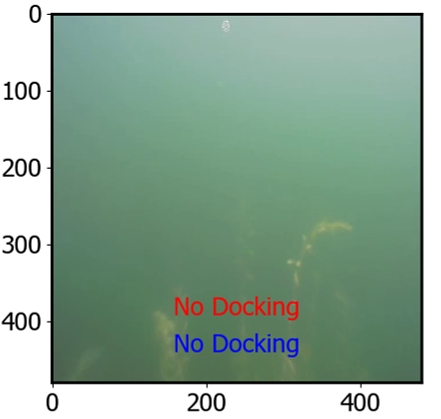}
    \caption{}
    \label{fig:qual_nd_res}
\end{subfigure}
\begin{subfigure}{0.42\columnwidth}
    \includegraphics[width=\textwidth]{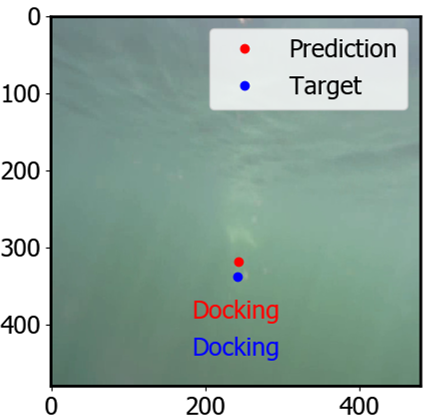}
    \caption{}
    \label{fig:qual_d_pyr}
\end{subfigure}
\hfill
\begin{subfigure}{0.42\columnwidth}
    \includegraphics[width=\textwidth]{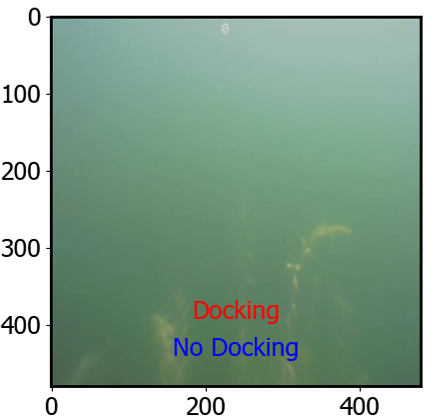}
    \caption{}
    \label{fig:qual_nd_pyr}
\end{subfigure}
\begin{subfigure}{0.42\columnwidth}
    \includegraphics[width=\textwidth]{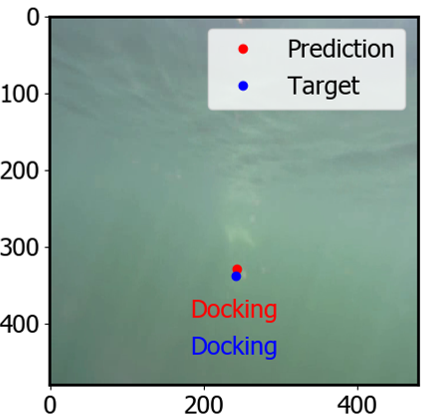}
    \caption{}
    \label{fig:qual_d_pyr_res}
\end{subfigure}
\hfill
\begin{subfigure}{0.42\columnwidth}
    \includegraphics[width=\textwidth]{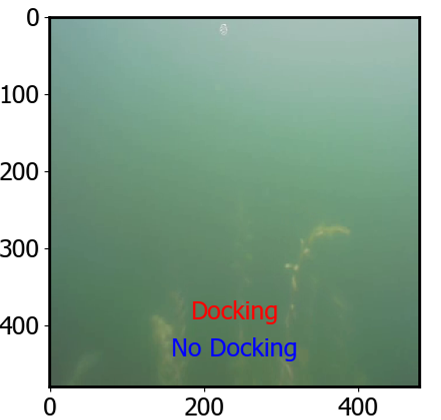}
    \caption{}
    \label{fig:qual_nd_pyr_res}
\end{subfigure}
\begin{subfigure}{0.42\columnwidth}
    \includegraphics[width=\textwidth]{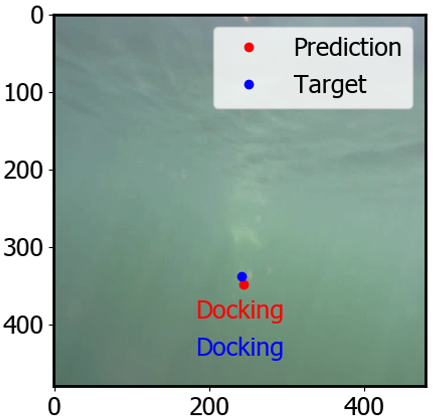}
    \caption{}
    \label{fig:qual_d_trans}
\end{subfigure}
\hfill
\begin{subfigure}{0.42\columnwidth}
    \includegraphics[width=\textwidth]{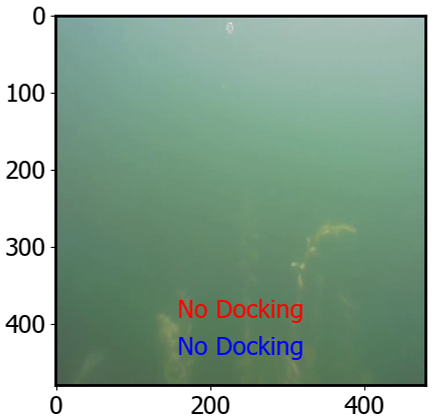}
    \caption{}
    \label{fig:qual_nd_trans}
\end{subfigure}
\caption{
(a),(c),(e),(g), and (i) show the prediction vs the actual docking position for the CNN, Residual, Pyramidal, Pyramidal-Residual, and Transformer architectures respectively. While (b),(d),(f),(h), and (j) show the predicted class when no docking station is present for the CNN, Residual, Pyramidal, Pyramidal-Residual, and Transformer architectures respectively.}
\label{fig:qualitative_detection}
\end{figure}

The qualitative analysis presented in Fig.~\ref{fig:qualitative_detection} contains one image (left side) with the docking station present, some noise in the form of glare and surface reflections. It also contains a second image (right side) without a docking station present, but with algae that create some errors in the classification output of the CNN, pyramidal, and pyramidal residual networks. In general, the classification output when the docking station is present is good, but we can observe some errors when the docking station is not present. This behavior supports the analysis of the loss for the test and train datasets, where the residual network exhibited the best performance overall. 

This can be explained by the ability of the residual blocks to allow a more efficient path for the data when necessary, thus allowing a direct connection from the input to the fully connected network at the network's output. As a result, this gives more granularity to determine the light's position and better accuracy in determining if the docking station is present. A summary of the results and a comparison of the performance of all the networks are shown in Table~\ref{tbl:archs_metrics}; the inference time presented was obtained after implementing each of the networks on the Nvidia Jetson Nano.

\begin{table}[H]
\centering
\include{table_archs.tex}
\caption{Performance of the different architectures on $X^{test}$.
\label{tbl:archs_metrics}}
\end{table}
\vspace{-0.7cm}
\subsubsection{Knowledge Distillation}
After finding that the residual network was the best-performing network, a student network was trained using knowledge distillation under the teacher-student framework for 60 epochs. This reduced the network footprint and inference time. The compressed network achieved an inference time of 71.4133 ms, using up to 183 MB of memory. The final L1 loss and accuracy were 0.0074 and 0.9907, respectively, which shows that the method improves the speed by a factor of 43\%, sacrificing only 0.39\% in docking localization and 0.59\% in accuracy.

The compressed architecture was integrated into the overall docking system and evaluated based on the docking success rate first in simulation and then during field deployments. It is important to mention that almost any of the proposed networks whose inference time is under 200 ms could have been used for the detection task since localization error is under 1\% for all of them; however, we decided to use the residual student network since it has the lowest memory footprint and offers a comparable inference time to the fastest (CNN).

\subsection{Underwater Simulation}
\begin{figure}[H]
\centering
\includegraphics[width=0.48\textwidth]{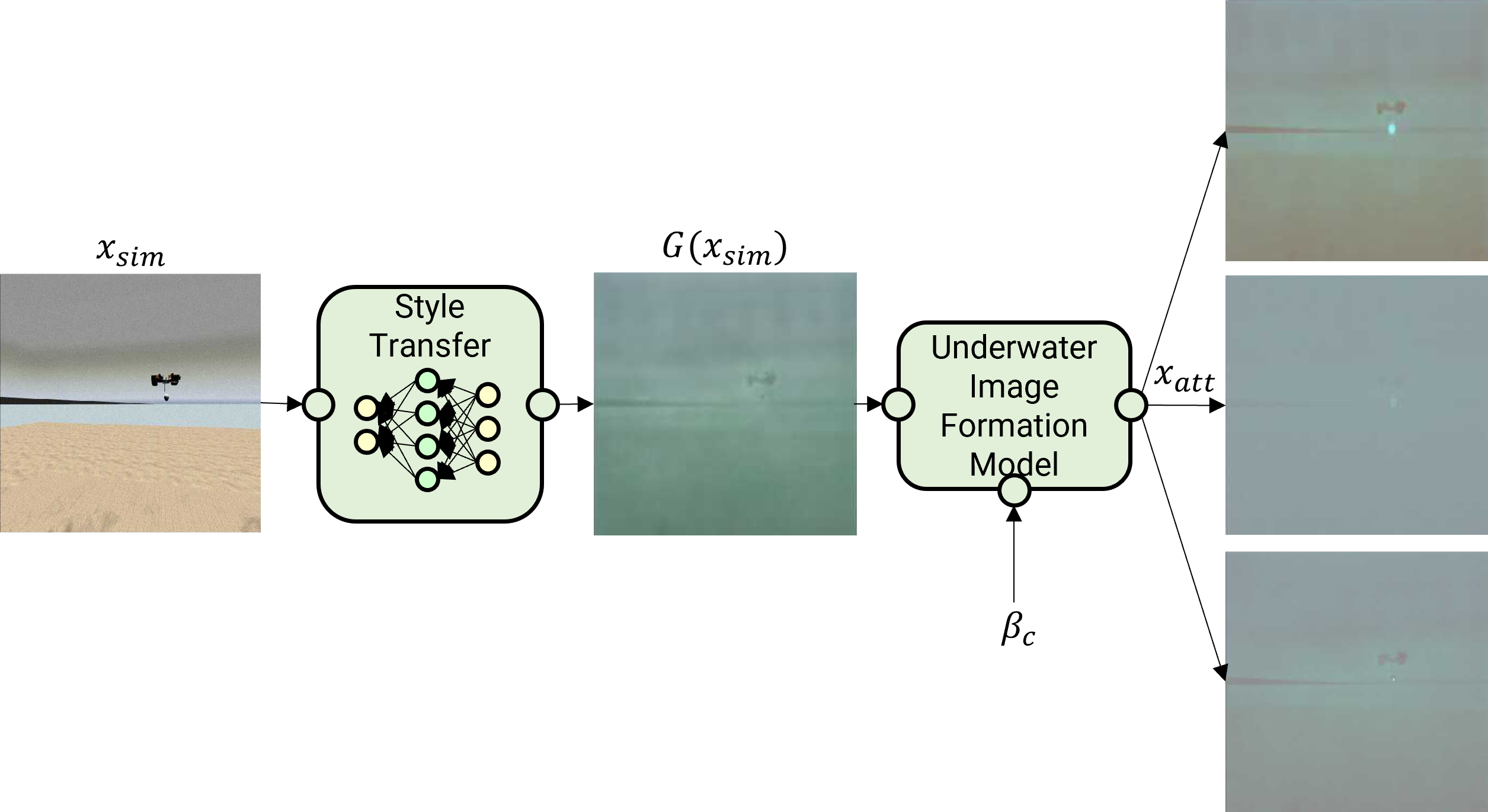}
\caption{The diagram shows on the left the image acquired directly from the simulated camera, in the middle the image after applying the learned style transfer, and on the right the possible attenuated images after applying the underwater image formation models with different attenuation coefficients $\beta_c$.}
\label{fig:pix2pix_uw}
\end{figure}

In order to assess the performance of each of these networks in the docking scenario, we implemented a style transfer approach that gave an underwater-like appearance to the images acquired by the camera in the Gazebo simulation. The style transfer network was based on the Pix2Pix implementation reported in \cite{isola2017image}, with an additional cycle consistency loss as the one reported on \cite{zhu2017unpaired}. The modified Pix2Pix network was trained for $500$ epochs. The results obtained after training this network can be observed in Fig.~\ref{fig:pix2pix_uw}.

Once the network was trained, it was integrated as part of the simulation, in such a way that the detection networks used the underwater-looking image instead of directly using the simulated camera output.

A test bench was created in the Gazebo simulation to assess the selected detection architecture's capability to guide the AUV successfully to the docking envelope. The test bench concluded when the AUV reached 20 successful docking attempts. At the beginning of each attempt, the vehicle is spawned at a random location 10 m - 15 m in front of the docking station and given a random heading. We accounted for an error in acoustic localization, modeled as $\pmb{\epsilon}\sim\mathcal{N}(0,\pmb{\sigma}^2_{dock})$, with $\pmb{\sigma}=[2m,2m,5^{\circ}]$.

To provide a reference for comparing our neural network (NN) based approach, a second method based on detecting the brightest pixel (BP) in the image was implemented. A total of thirteen tests were conducted, assessing the success rate of the two different detection algorithms under three distinct underwater current conditions (0.25 m/s, 0.1 m/s, 0 m/s) perpendicular to the docking direction, and two different Jerlov water types, IC (low turbidity), and 5C (high turbidity). To further study impact of water turbidity in the docking success rate, an additional test case was created with the Jerlov water type 7C and a random direction underwater current of 0.05 m/s. The results of our experiments are summarized in Table~\ref{tbl:table_docking}. Overall, our NN based approach outperformed the BP method in 4 out of the 6 test scenarios, obtained the same result in one, and performed worse in one additional case. This is mainly due to the ability of the NN to provide a stable docking localization even in the presence of glare, or other noise. Additionally, the NN provides a classification output, which is used to turn on and off the optical guidance; this allows the overall algorithm to keep a defined target even when the docking station has not been detected. Another significant observation from our tests concerns the influence of water clarity versus underwater current. While disturbances negatively affected the success rate, as evident in T1.1-T1.6, a substantial improvement is notable after reducing water turbidity, with tests (T2.1-T2.6) achieving a success rate exceeding 90\%. This change in water turbidity modifies the detection range from $\sim 6$ m in the high turbidity case to $\sim 12$ m in the low turbidity case. In this series of tests (T2.1-T2.6), it appears that the underwater current has a minimal impact on success, which can be attributed to the detection algorithms' ability to function effectively from a distance, providing the AUV with the necessary time to reduce errors and align with the docking station centerline.

Test case 3.1 was designed to replicate conditions similar to those found in Fairfield Lakes, IN, the planned location for field deployments. In this scenario, the simulation was configured to mimic Jerlov water type 7C and a random underwater current of 0.05 m/s. The increased turbidity of the water reduced visibility to approximately 4 meters, which had an impact on the success rate compared to the other test cases. After running this test until ten successful docking attempts were achieved, the overall success rate reached 0.32. This outcome is reasonable, given that the terminal homing initially depends on the acoustic localization estimate. If there is an error in this estimate, the only way to correct it is through optical guidance. If the optical guidance is occluded until the very last part of the docking procedure, then the AUV doesn't have enough time to react and dock successfully.

\begin{table}[H]
\centering
\include{table_docking.tex}
\caption{Docking success rate under different environmental conditions, including high turbidity, low turbidity, and strong, mild, and no underwater current.
\label{tbl:table_docking}}
\end{table}

\subsection{Field Deployments}
Several tests were performed on the Fairfield Lakes in Lafayette, IN for the field deployments. The docking station was mounted on the back of the Boat for Robot Engineering and Applied Machine Learning (BREAM-ASV). The BREAM-ASV was anchored to the bottom of the lake in three points, allowing slow translational and rotational movement. This simulated the station keeping mechanism required for the docking procedure. The AUV was started in the vicinity of the BREAM-ASV to allow acoustic communication to be established quickly. Since the latch mechanism in the docking station is composed of permanent magnets, every time a successful docking occurred, the AUV had to be restarted and manually unlatched. Additionally, the anchors were dragged out of place during each successful docking or collision with the docking station, causing the BREAM-ASV to need re-anchoring. This made testing an involved process. 

\begin{figure*}[!th]
\centering
\begin{subfigure}{0.20\textwidth}
    \includegraphics[width=\textwidth, height=\textwidth]{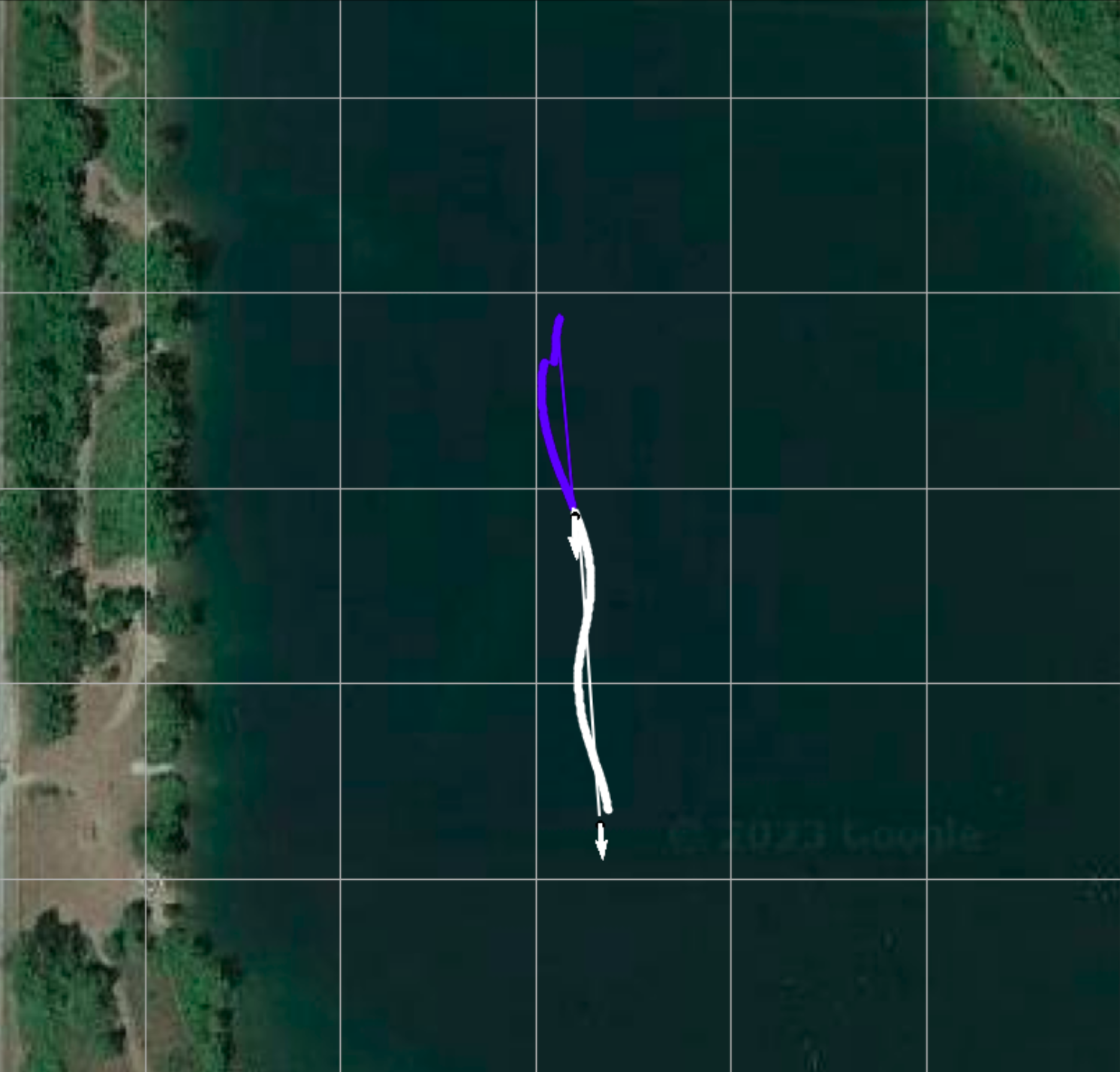}
    \caption{}
    \label{fig:failure_docking}
\end{subfigure}
\hfill
\begin{subfigure}{0.20\textwidth}
    \includegraphics[width=\textwidth, height=\textwidth]{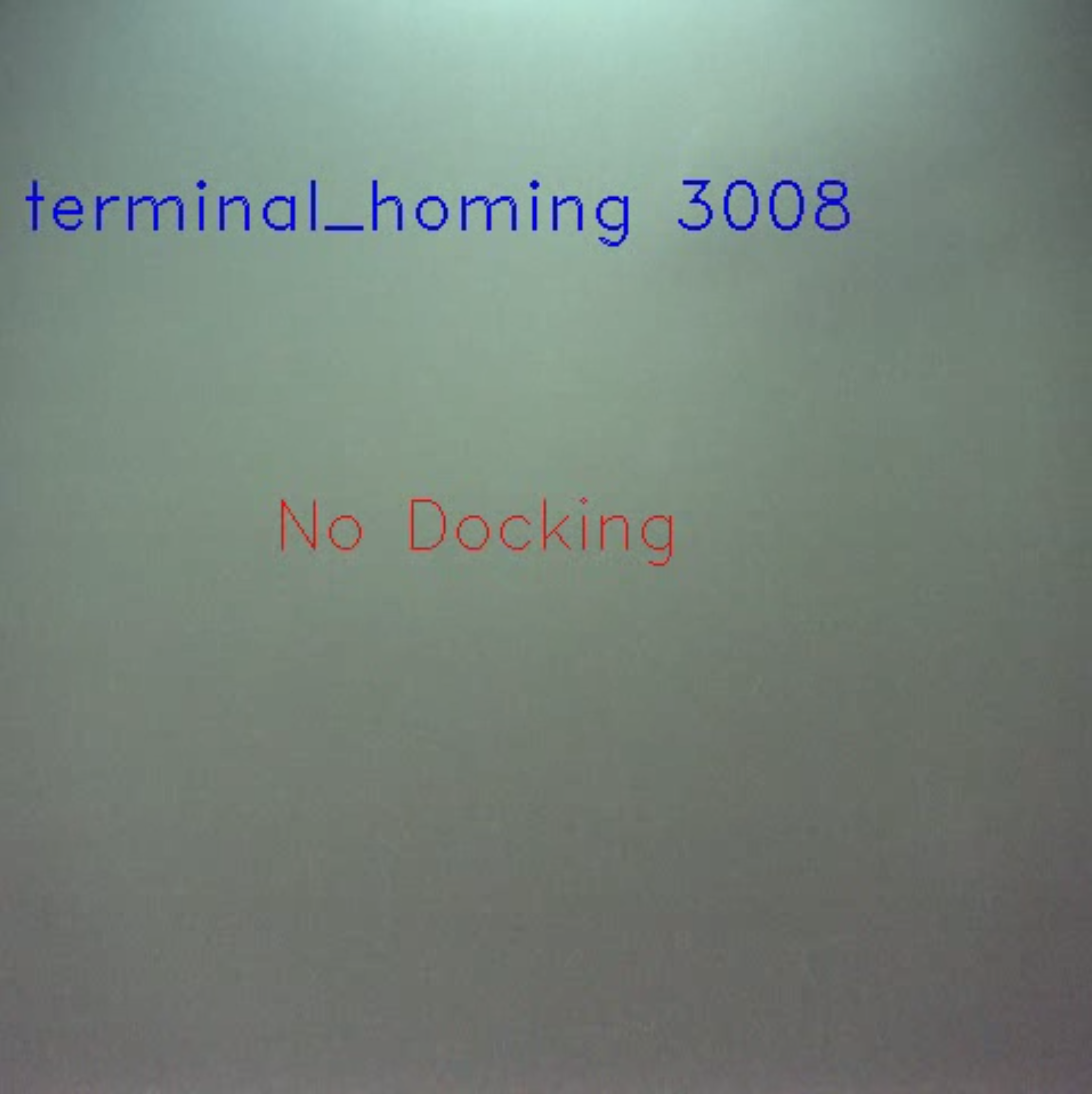}
    \caption{}
    \label{fig:failure_docking}
\end{subfigure}
\hfill
\begin{subfigure}{0.20\textwidth}
    \includegraphics[width=\textwidth, height=\textwidth]{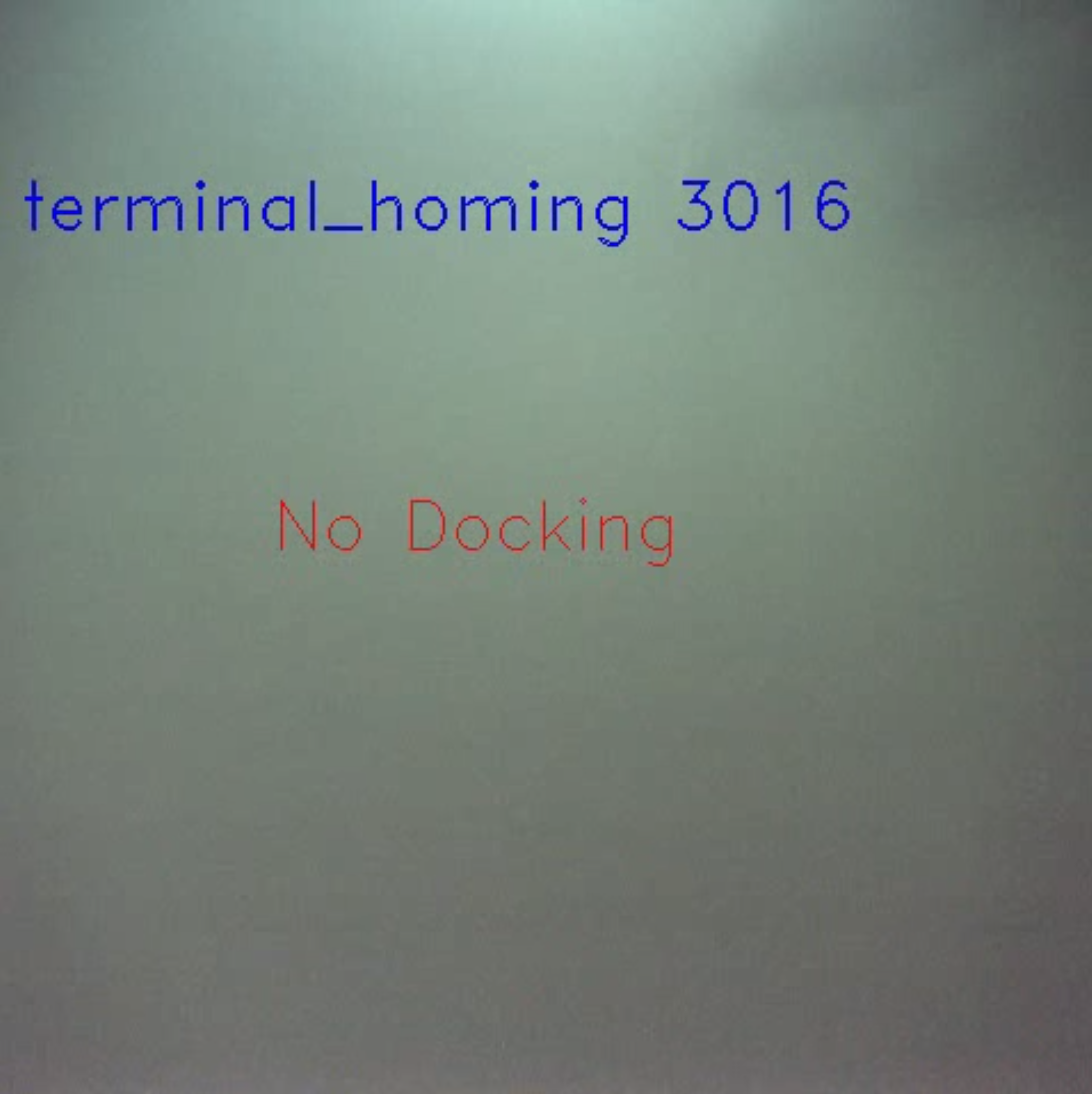}
    \caption{}
    \label{fig:failure_docking}
\end{subfigure}
\hfill
\begin{subfigure}{0.20\textwidth}
    \includegraphics[width=\textwidth, height=\textwidth]{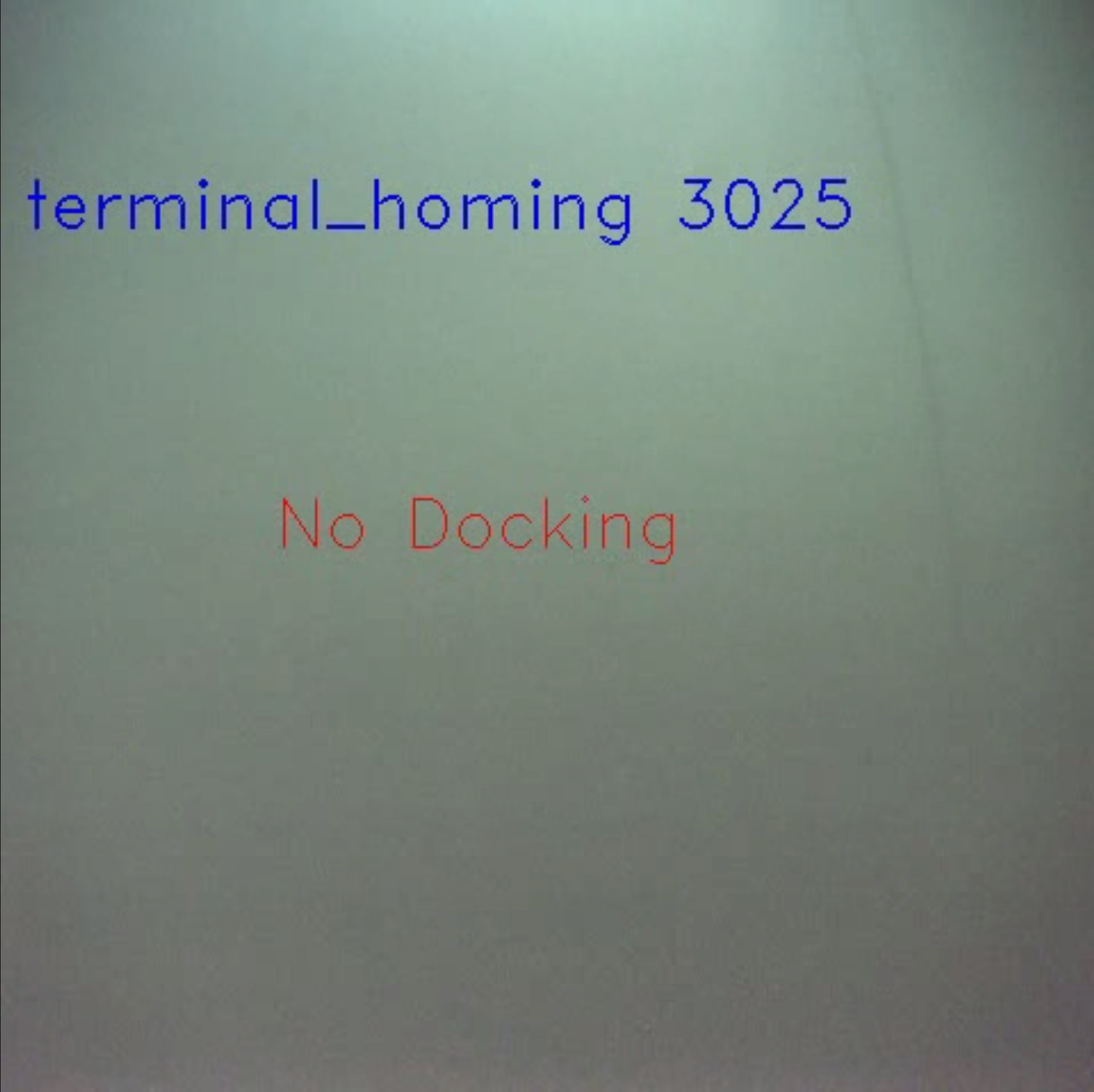}
    \caption{}
    \label{fig:failure_docking}
\end{subfigure}
\caption{An unsuccessful docking case in which the Iver3 AUV passes by the docking station at a distance, preventing the light from ever entering the camera frame. (a) shows the path taken in this case. On the plotted path, blue shows the approach phase, white shows the terminal homing phase, and pink represents the points at which the Iver3 AUV has detected a docking station from the camera. (b),(c), and (d) are a sequence of frames taken by the Iver3 AUV camera. In this sequence, the ASV and the docking station pass by in the top right corner of the camera frame.}
\label{fig:failed_docking_analysis}
\end{figure*}

\begin{figure*}[!th]
\centering
\begin{subfigure}{0.20\textwidth}
    \includegraphics[width=\textwidth, height=\textwidth]{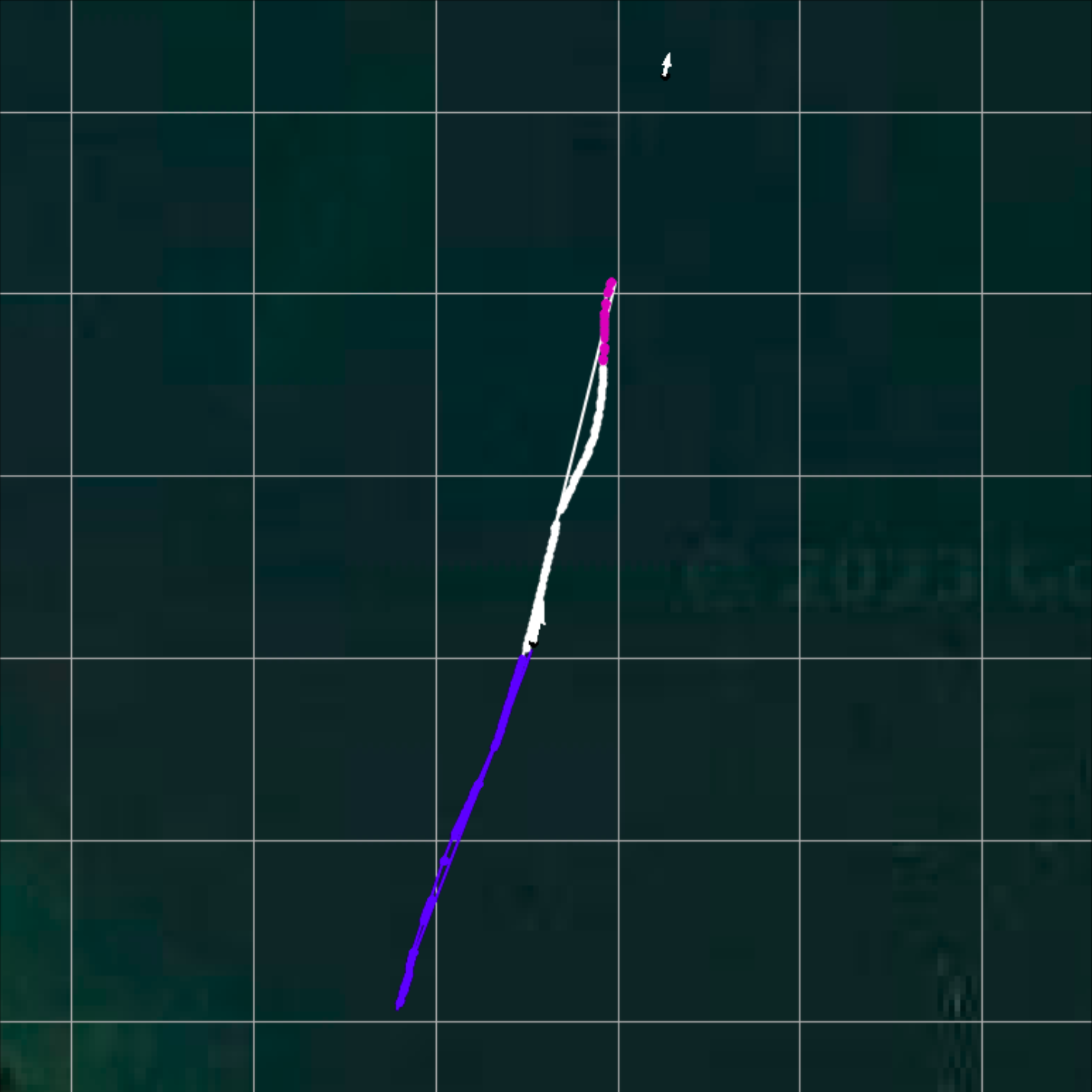}
    \caption{}
    \label{fig:successful_docking}
\end{subfigure}
\hfill
\begin{subfigure}{0.20\textwidth}
    \includegraphics[width=\textwidth, height=\textwidth]{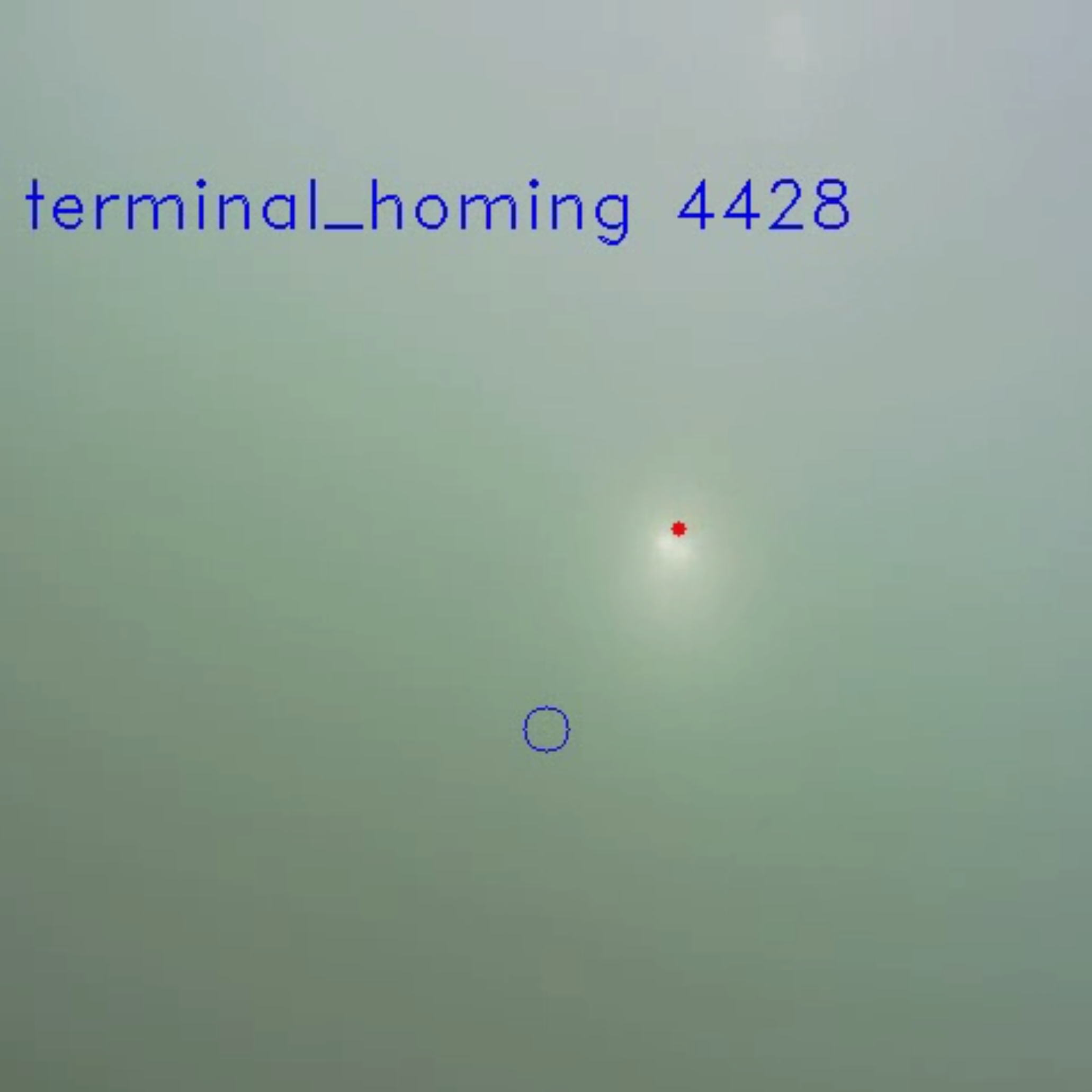}
    \caption{}
    \label{fig:successful_docking}
\end{subfigure}
\hfill
\begin{subfigure}{0.20\textwidth}
    \includegraphics[width=\textwidth, height=\textwidth]{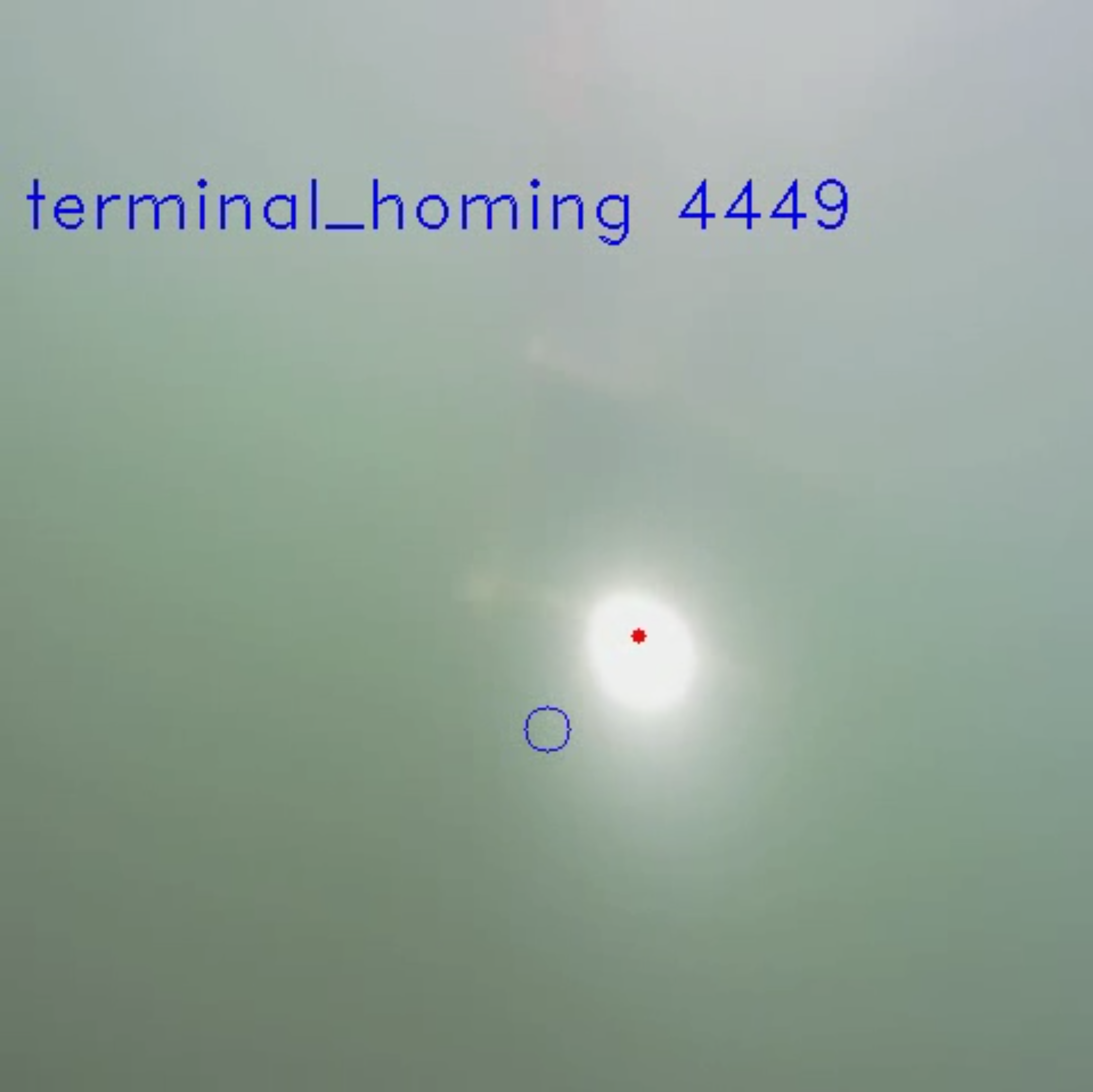}
    \caption{}
    \label{fig:successful_docking}
\end{subfigure}
\hfill
\begin{subfigure}{0.20\textwidth}
    \includegraphics[width=\textwidth, height=\textwidth]{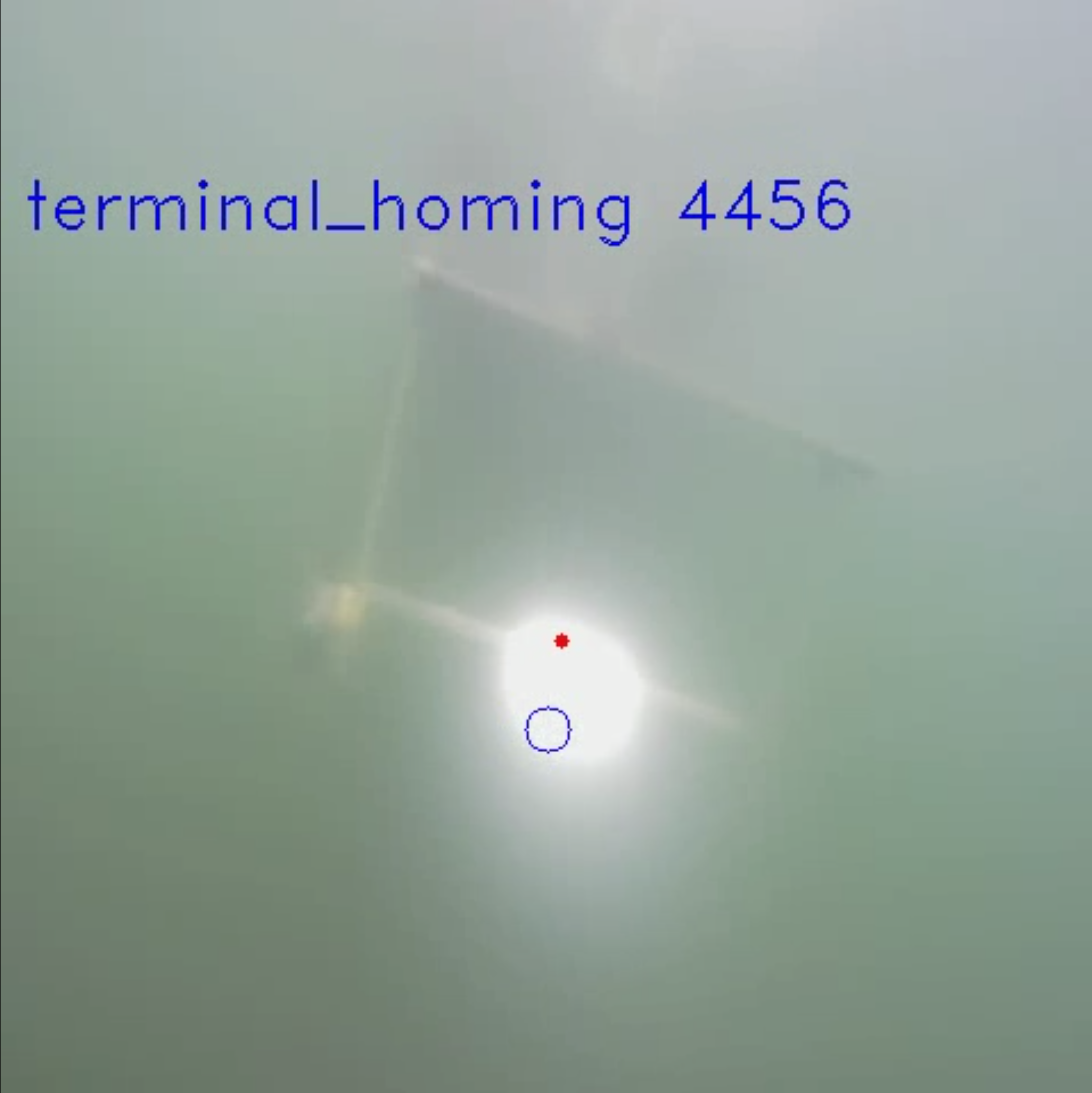}
    \caption{}
    \label{fig:successful_docking}
\end{subfigure}
\caption{A successful docking case. (a) shows the path taken in this case. On the plotted path, blue shows the approach phase, white shows the terminal homing phase, and pink represents the points at which the Iver3 AUV has detected a docking station from the camera. (b),(c), and (d) are a sequence of frames taken by the Iver3 AUV camera. These frames were selected from the end of the terminal homing stage. In these frames, the red dot is the detected location of the docking station and the blue circle is a reference for the horizontal midpoint of the frame.}
\label{fig:successful_docking_analysis}
\end{figure*}

A total of 50 tests were executed over 4 different days, with a total of 16 successful docking attempts. This success rate is acceptable for underwater docking because the retry mechanism allows the AUV to replan and make another docking attempt after a docking failure. Since each retry attempt takes about three minutes to complete, and with each attempt, the AUV gains a better localization estimate, the retry mechanism is a low-risk maneuver. 
Some attempts were unsuccessful due to low visibility and heading error due to cross track oscillations, an example can be seen in Fig.~\ref{fig:failed_docking_analysis}. This caused the docking station to not be fully visible by the AUV, and in other cases, caused the docking station to be visible only after it was too late for the AUV to react. The minimum necessary detection distance for docking to be possible for this procedure is around 4m. With a maximum visibility of around 8m, if the vehicle approached the docking station at an angle due to oscillations and/or the localization error was too high, the vehicle would pass by the docking station without it entering the camera frame. The large cross track oscillations were addressed after the third test day, when the ILOS controller was re-tuned to mitigate the oscillations in the path following. As a result, on the fourth test day, the docking station was in the image frame for longer, giving the neural network more time to guide the AUV than in previous testing days. This slight improvement can be observed in Fig.~\ref{fig:successful_docking_analysis}, which presents a successful docking with smaller oscillations than in the failed docking attempt in Fig.~\ref{fig:failed_docking_analysis}. Since the latch mechanism in the docking station is composed of permanent magnets, each time a successful docking occurred, the AUV had to be restarted and manually unlatched. Additionally, the anchors were dragged out of place during each successful docking or collision with the docking station, causing the BREAM-ASV to need re-anchoring. This made testing an involved process; however, with an ASV capable of station-keeping, the human intervention will not be needed. 

\section{Conclusions}\label{sec:conc}

In conclusion, this research addresses the problem of underwater docking for AUVs by employing deep learning based detection architectures. To validate and understand the challenges associated to the docking procedure, a simulation was developed based on Gazebo and ROS. The simulation included the AUV dynamics, the errors associated to the acoustic localization, and the visibility problem when dealing with underwater vision. This allowed a straightforward implementation into the AUV Iver3 platform.

We compared various deep learning architectures commonly used for vision tasks such as CNN, Residual, Pyramidal, Pyramidal Residual, and Transformers. The different architectures were trained for the docking detection and classification task. After the training, we found that all the networks exhibited a localization error below 1\% of the image width, and a classification accuracy above 98\%. From all the networks, the residual exhibited the best performance with a localization error of 0.3402\% and accuracy of 99.66\%. To facilitate the real time implementation, knowledge distillation was employed to compress the residual network and increase its inference time. This led to a final compressed residual network with a localization error of 0.7360\%, accuracy of 99.07\%, an inference time of 71.14ms and a reduction in the number of parameters of 60\%.

To bridge the simulation to reality gap, we developed an acoustic simulation that allowed localization errors, as well as communication failures. Additionally, we used a Generative Adversarial Network (GAN) for image-to-image translation, transforming Gazebo simulation images into realistic underwater looking images. These images were then processed using underwater image formation models to simulate image attenuation under different water conditions and over varying distances. All these components allowed to tune the docking strategy to different environmental conditions, and also to understand its impact on the docking success rate.

The proposed approach was evaluated in simulation and in the field. The results in simulation demonstrated an improvement of at least 20\% in docking success rate over classical vision methods under challenging visibility conditions. Furthermore, the approach's effectiveness was validated through experimental results using the off-the-shelf platform Iver3. A total of 66 tests were executed in 4 different days, with a total of 16 successful docking attempts and 50, giving an an overall success rate of 0.32. This result is the same as the one obtained in the extreme simulation scenario with the Jerlov water type 7C, which supports the idea that the docking success rate is highly tied to visibility. Although the success rate is not perfect, the retry mechanism implemented as part of the docking strategy adds to the robustness of the approach, which makes it a viable solution even in low visibility scenarios.

Future development of this work includes refinement to the detection algorithm using recurrent networks. This would help to capture a temporal component in the detection process, thus improving the robustness and increasing the range of detection in cases of low visibility. Additionally, further refinement of the light beacon and the docking station's mechanical design can be done to gain visibility range. Finally, integrating the docking strategy with an actively controlled ASV would reduce the relative heading error between the docking station and the AUV, thus improving the overall success rate.

\bibliographystyle{IEEEtran}
\bibliography{references.bib}

\end{document}

%% file: table_archs.tex
\begin{tabular}{ccccc}
\cline{2-5}
                                                             & L1 Loss & Accuracy & \begin{tabular}[c]{@{}c@{}}Inference\\ Time (ms)\end{tabular} & \begin{tabular}[c]{@{}c@{}}Number of \\ Parameters\end{tabular} \\ \hline
CNN                    & 0.004711         & 0.994078         & 66.5526         & 28,263,716\\
Residual               & 0.003402         & 0.996616         & 102.0481        & 39,102,980\\
Pyramidal              & 0.008639         & 0.98308          & 137.9272        & 73,774,020\\
\begin{tabular}[c]{@{}c@{}}Pyramidal\\ Residual\end{tabular}
                       & 0.007176         & 0.992386         & 134.6221        & 74,480,196\\
Transformer            & 0.008955         & 0.988946         & 458.3791        & 29,866,628\\ \hline
\end{tabular}

%% file: table_docking.tex
\renewcommand{\tabcolsep}{1.5pt}

\begin{tabular}{cccccc}
\cline{2-6}
         & \begin{tabular}[c]{@{}c@{}}Detection\\ Method\end{tabular} & \begin{tabular}[c]{@{}c@{}}Current\\ (m/s)\end{tabular} & \begin{tabular}[c]{@{}c@{}}Water\\ Trubidity\end{tabular} & \begin{tabular}[c]{@{}c@{}}Docking \\ Success \\ Rate\end{tabular} & \begin{tabular}[c]{@{}c@{}}Cross \\ Track\\ Error (m)\end{tabular} \\ \hline
Test 1.1 & NN                                                         & 0.0                                                     & High                                                      & 0.9524                                                             & 0.5498                                                             \\
Test 1.2 & BP                                                         & 0.0                                                     & High                                                      & 0.7407                                                             & 0.5478                                                             \\
Test 1.3 & NN                                                         & 0.1                                                     & High                                                      & 0.8333                                                             & 0.6244                                                             \\
Test 1.4 & BP                                                         & 0.1                                                     & High                                                      & 0.6250                                                             & 0.6244                                                             \\
Test 1.5 & NN                                                         & 0.25                                                    & High                                                      & 0.7407                                                             & 0.7148                                                             \\
Test 1.6 & BP                                                         & 0.25                                                    & High                                                      & 0.4651                                                             & 0.6960                                                             \\
Test 2.1 & NN                                                         & 0.0                                                     & Low                                                       & 1                                                                  & 0.5710                                                             \\
Test 2.2 & BP                                                         & 0.0                                                     & Low                                                       & 0.9090                                                             & 0.5526                                                             \\
Test 2.3 & NN                                                         & 0.1                                                     & Low                                                       & 0.9090                                                             & 0.6523                                                             \\
Test 2.4 & BP                                                         & 0.1                                                     & Low                                                       & 0.9523                                                             & 0.7147                                                             \\
Test 2.5 & NN                                                         & 0.25                                                    & Low                                                       & 1                                                                  & 0.7237                                                             \\
Test 2.6 & BP                                                         & 0.25                                                    & Low                                                       & 1                                                                  & 0.7010                                                             \\ \hline
Test 3.1 & NN                                                         & 0.05                                                    & Very High                                                       & 0.32                                                                  & 0.7333                                                             \\ \hline
\end{tabular}